\pdfoutput=1

\documentclass[11pt]{article}

\usepackage{emnlp2021}

\usepackage{times}
\usepackage{latexsym}

\usepackage[T1]{fontenc}

\usepackage[utf8]{inputenc}

\usepackage{microtype}

\usepackage{graphics}
\usepackage{graphicx}

\usepackage{amsthm}
\usepackage{amsmath}
\usepackage{amssymb}
\usepackage{multirow}
\usepackage{amsfonts}
\usepackage[normalem]{ulem}

\usepackage{booktabs}
\usepackage[]{caption}
\DeclareCaptionLabelFormat{AppendixTables}{A.#2}

\newcommand{\av}{\mathbf{a}}
\newcommand{\bv}{\mathbf{b}}
\newcommand{\cv}{\mathbf{c}}

\newcommand{\rv}{\mathbf{r}}

\newcommand{\xv}{\mathbf{x}}
\newcommand{\yv}{\mathbf{y}}

\theoremstyle{definition}
\newtheorem{definition}{Definition}[section]

\newcommand{\mingkai}[1]{#1}

\title{Compression, Transduction, and Creation: \\
A Unified Framework for Evaluating Natural Language Generation}

\author{
Mingkai Deng$^{1*}$,~~
Bowen Tan$^{1*}$,~~
Zhengzhong Liu$^{1,2}$,~~
Eric P. Xing$^{1,2,3}$,~~
Zhiting Hu$^{4}$\\
$^1$Carnegie Mellon University,~~ $^2$Petuum Inc.,~~ $^3$MBZUAI,~~ $^4$UC San Diego\\
($^*$equal contribution) \\
{\small 
{\tt \{mingkaid,btan2,liu,epxing\}@andrew.cmu.edu, zhh019@ucsd.edu}
}
}

\begin{document}

\maketitle


\begin{abstract}
Natural language generation (NLG) spans a broad range of tasks, each of which serves for specific objectives and desires different properties of generated text. The complexity makes automatic evaluation of NLG particularly challenging. Previous work has typically focused on a single task and developed individual evaluation metrics based on specific intuitions. In this paper, we propose a unifying perspective \mingkai{that facilitates the design of metrics for a wide range of language generation tasks and quality aspects.} 
Based on the nature of information change \mingkai{from input to output, we classify NLG tasks into} compression (e.g., summarization), transduction (e.g., text rewriting), and creation (e.g., dialog). The \emph{information alignment}, or overlap, between input, context, and output text
plays a common central role in characterizing the generation. \mingkai{Using the uniform concept of information alignment}, we develop a family of interpretable metrics for various NLG tasks and aspects, often without need of gold reference data. 
To operationalize the metrics, we train self-supervised models to approximate information alignment as a prediction task. 
Experiments show the uniformly designed metrics achieve stronger or comparable correlations with human judgement compared to state-of-the-art metrics in each of diverse tasks, including text summarization, style transfer, and knowledge-grounded dialog. \mingkai{With information alignment as the \textit{intermediate representation}, we deliver a composable library for easy NLG evaluation and future metric design.}\footnote{Code available at \url{https://github.com/tanyuqian/ctc-gen-eval}}

\end{abstract}

\begin{figure}[t]
    \centering
    \includegraphics[width=0.45\textwidth]{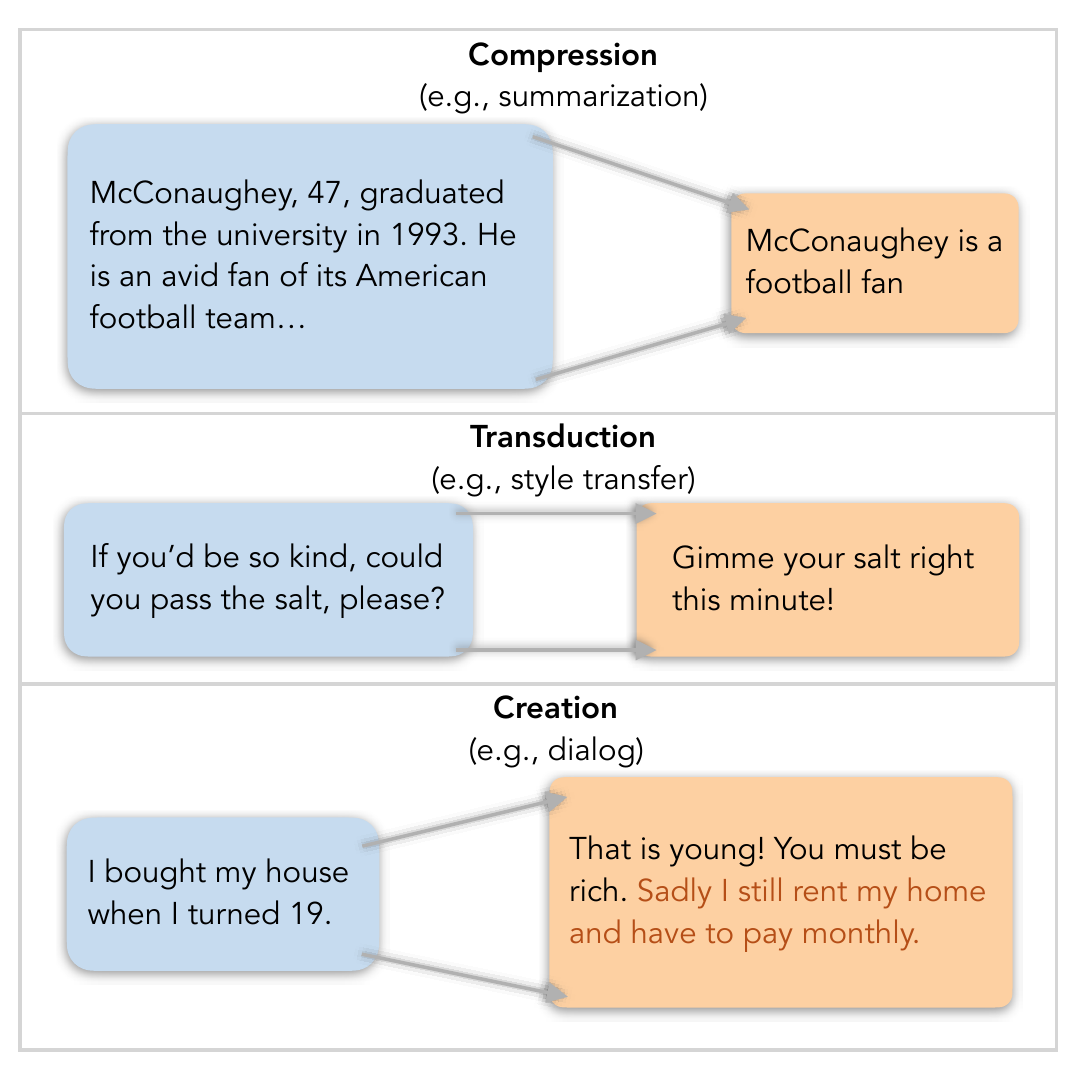}
    \vspace{-10pt}
    \caption{
    Illustration of three categories of NLG tasks in terms of information change. Task input is in {\color{blue} blue} box and output in {\color{orange} orange} box. Text in {\color{red} red} in the dialog output box represents newly created information.
    }
    \label{fig:framework-illustration}
\end{figure}

\section{Introduction}\label{sec:intro}

Natural language generation (NLG) refers to the broad set of tasks that produce fluent text from input data and other contextual information. The diverse tasks serve for vastly different uses in practice. For example, \emph{summarization} compresses a source article into a short paragraph containing the most important information; \emph{translation} transduces content expressed in one language into another; and a \emph{chatbot} creates novel responses to drive the conversation. Recent years have seen remarkably fast progress in improving and making new models for NLG tasks. However, evaluation of NLG has long been considered difficult \cite{kryscinski-etal-2019-neural,mathur2020tangled}: human evaluation is often prohibitively expensive and slow, while accurate automatic evaluation is challenging given the complexity of text modeling and the diverse aspects to be measured for different NLG tasks.

Previous work has developed a large variety of automatic metrics. A popular general strategy is to measure the similarity of generated text against human-written references, such as the classical BLEU \cite{papineni-etal-2002-bleu}, ROUGE \cite{lin-2004-rouge}, and more recent variants based on neural models \cite[e.g.,][]{zhang2020bertscore,sellam-etal-2020-bleurt}. However, an NLG task typically involves multiple desirable properties (e.g., consistency, conciseness, richness) that may have different priorities and need trade-off depending on the application scenarios \cite{hashimoto2019unifying,mir-etal-2019-evaluating,mehri2020usr,gehrmann2021gem}.
Thus a single score without multi-aspect interpretability is often inadequate to characterize generation quality. A growing number of recent works have proposed aspect-based metrics for popular tasks such as summarization \cite{kryscinski2019evaluating,wang2020asking} and dialog \cite{mehri2020usr,nie2020i}. Those metrics are typically each designed for individual tasks and aspects, based on specific intuitions. The lack of a common theoretical ground makes it difficult to share the evaluation strengths across the diverse NLG problems, and fails to offer guidance to metric design for emerging tasks and aspects.

In this paper, we propose a more unifying perspective of NLG evaluation through the lens of information change, which offers a general framework to measure many key aspects of NLG tasks. In particular, based on the practical use of NLG, each task can be seen as one of (1) \emph{compression} to express salient information in concise text, such as summarization and image captioning; (2) \emph{transduction} to transform text while preserving content precisely, such as translation and style transfer; and (3) \emph{creation} to produce new content from input context, such as dialog and story generation. A common concept underlying the three broad categories is \emph{information alignment}, which we define as the extent to which the information in one generation component is grounded in another. Here the generation components include input, output, additional context, and references when available. 

Inspired by recent work on model-based evaluation, we adopt contextualized language models to measure information alignment. 
We then demonstrate the framework by devising a family of highly intuitive metrics for three representative tasks (\textit{aspects}) in each category \mingkai{uniformly in terms of information alignment}, including summarization (\textit{relevance} and \textit{consistency}), style transfer (\textit{content preservation}) and knowledge-based dialog (\textit{engagingness} and \textit{groundedness}), respectively. 
Experiments show that the uniformly designed metrics robustly outperform or compete with state-of-the-art metrics specifically designed for each task, 
in terms of correlations with human judgement.
We also study different implementations of the central information alignment estimation model, showing that improved alignment measure leads to better evaluation quality across all the tasks/aspects.

\section{Related Work}


\paragraph{Task- and Aspect-Specific NLG Evaluation.} 
Canonical automatic evaluation~\cite{papineni-etal-2002-bleu,lin-2004-rouge} often compute a single score measuring some forms of similarity between outputs and human-written references. The later-emerged learning-based approaches aggregate multiple features to regress on human-rated quality scores for different tasks~\cite{lowe-etal-2017-towards,peyrard-etal-2017-learning,sellam-etal-2020-bleurt}. Researchers also identified that a single evaluation score cannot account for the variety of quality factors that exist in multi-faceted NLG applications. A number of metrics were then proposed for specific tasks, either to evaluate multiple aspects~\cite{mehri2020usr,egan2021play} or to focus on one particular aspect~\cite{kryscinski2019evaluating,mehri2020unsupervised,nie2020i,Durmus_2020,wang2020asking}. Our framework continues this line of research to produce interpretable metrics for multiple aspects. While recent evaluation frameworks each discussed the key evaluation aspects of one NLG task~\cite{venkatesh2018evaluating,mir-etal-2019-evaluating,yamshchikov2020styletransfer,fabbri2021summeval}, our framework provides a unified methodology that facilitates metric design for all the three main categories of tasks. We also highlight that all of metrics (except for the relevance metric for summarization) are reference-free once trained.

Several emerging NLG benchmarks~\cite{gehrmann2021gem,liu-etal-2021-glge} collected existing metrics for various tasks, 
whereas we aim at developing new unified metrics with stronger performance. \citet{belz-etal-2020-disentangling} proposed a categorization for different NLG quality aspects. Our general framework covers all the described types of quality.

\paragraph{Text-to-Text Information Alignment.} 
Measuring information overlap between texts is a recurring theme in designing NLG evaluation metrics. It has typically been approximated by n-gram overlap~\cite{papineni-etal-2002-bleu,popovic-2015-chrf}, 
synonym matching~\cite{banerjee-lavie-2005-meteor} and 
embedding similarities~\cite{kusner2015word}. 
Recently, pre-trained models~\cite{devlin-etal-2019-bert} were introduced to improve token-level embedding matching~\cite{zhang2020bertscore} and leverage extrinsic capabilities such as question answering \cite{eyal-etal-2019-question,wang2020asking} and entailment classification \cite{falke-etal-2019-ranking,kryscinski2019evaluating,zhou2020detecting} to align variable spans and entire sentences.
\citet{egan2021play} proposed automatic Shannon Game \cite{hovy-1998-shannon} to measure the decrease of the information one can gain from a document after observing its summary; \citet{peyrard-2019-simple} conducted a theoretical analysis to characterize the information change among source document, background knowledge and summaries. These methods are often restricted to a single task, while we offer a general framework adaptable to a wide range of tasks and aspects.

\section{A Unified Evaluation Framework}

We present the new framework that offers a common foundation for characterizing diverse NLG tasks and leads to a set of interpretable metrics for evaluating their key aspects. 

As discussed in \S\ref{sec:intro}, NLG tasks can be categorized as performing compression, transduction, or creation based on changes in conveyed information from input to output. For a {\bf compression} task (e.g., summarization), the goal is to concisely describe the most important information in the input (e.g., a document). That is, the output should only contain content from the input, namely \emph{``consistency''} \cite{cao2018faithful,kryscinski-etal-2019-neural,zopf-etal-2016-beyond,peyrard-2019-simple}, and the included content must be salient, namely \emph{``relevance''} \cite{nenkova-passonneau-2004-evaluating,zopf-etal-2016-beyond}. Intuitively, with an ``information alignment'' measure that assesses how the information in a generated output overlaps with that in the  input (and in references that offer clues for salience), we can readily evaluate the two key aspects. The same intuition applies to {\bf transduction} tasks (e.g., style transfer), where the output must preserve the input content precisely. The evaluation of \emph{``preservation''} \cite{mir-etal-2019-evaluating} thus also boils down to measuring the information alignment between input and output. A {\bf creation} task (e.g., dialog) generates output that adds on top of input (e.g., dialog history) new information (e.g., from external knowledge). Information alignment between the output, input, and external sources is thus essential for evaluating how well the created content \emph{engages} with the context \cite{venkatesh2018evaluating,see2019makes} and how meaningful the content is by \emph{grounding} to the external sources \cite{dinan2019second,smith-etal-2020-put}.

From the above perspective, information alignment arises as a common central component that connects evaluations across the tasks. A single accurate alignment prediction model would enable us to reliably evaluate many relevant aspects in various applications. 

Next, we first present our definition of information alignment (\S\ref{sec:notation}); then describe the details of how the aspect metrics for compression, transduction, and creation are built on the alignment (\S\ref{sec:compression}-\ref{sec:creation}); we finally discuss different effective implementations of the underlying alignment estimation model based on neural networks (\S\ref{sec:alignment_implementations}).

\subsection{Preliminaries}\label{sec:notation}
For an NLG task, 
let $\xv$ be the input, $\cv$ be any other additional context, and $\yv$ be the output text generated conditioning on $\xv$ and $\cv$. For example, in knowledge-based dialog, $\xv$ is the dialog history, $\cv$ is external knowledge such as a Wikipedia article, and $\yv$ is the response.
In the current work, we assume both $\xv$ and $\cv$ to be text, but the general framework is also applicable when $\xv$ and $\cv$ are in other modalities (e.g., images, tables), as long as we can measure their information alignment with $\yv$  as defined below (e.g. using cross-modal models).
In some tasks, gold standard output written by human is available, which we denote as $\rv$.

As above, information alignment is the central module for NLG evaluation. We consider the alignment from arbitrary text $\av$ to $\bv$ as token-level soft alignment. More formally:
\begin{definition}[Information Alignment]
\label{def:info-align}
Let $\av$ be a piece of text of length $N$; $\bv$ be arbitrary data. The information alignment from text $\av$ to $\bv$ is a vector of alignment scores:
\vspace{-5pt}
\begin{equation}
\small
    \textit{align}(\av \to \bv) = \langle \alpha_1, \alpha_2, \dots, \alpha_N \rangle,
\label{eq:alignment-def}
\end{equation}
where $\alpha_n \in [0,1]$ is the confidence that the information of the $n$-th token in $\av$ is grounded by $\bv$, i.e., the $n$-th token aligns with $\bv$.
\end{definition}
Note that the alignment is ``one-directional'' from $\av$ to $\bv$: it does not measure how $\bv$ aligns to $\av$. 
We next show how the alignment scores can be used to define intuitive metrics for various tasks. Besides, the fine-grained alignment scores also offer a certain level of interpretability for the resulting metrics, as illustrated by the example in Table~\ref{table:alignment-example}.

\subsection{Evaluation of ``Compression'' Tasks} \label{sec:compression}
We discuss compression evaluation in the context of text summarization, an extensively studied task for evaluation in previous work. The task aims to extract the most important information from document $\xv$ and express it in summary $\yv$. As above, \emph{consistency} and \emph{relevance} have been widely identified as key aspects to characterize the content quality of generated summaries \cite{cao2018faithful,kryscinski-etal-2019-neural,zopf-etal-2016-beyond,peyrard-2019-simple}. We propose our metrics below. 

\paragraph{Consistency} 
We adopt the prevailing definition of consistency \cite{cao2018faithful,kryscinski-etal-2019-neural}, which dictates that the summary $\yv$ should only contain information from $\xv$ (instead of other sources or hallucinations). The aspect is also referred to as ``factual correctness'' or  ``faithfulness'' in previous work\footnote{For the aspects studied in this paper, we summarize in Table \ref{table:task-aspects} the alternative names that used in previous work.}.
For $\yv$ to be fully consistent, all tokens in $\yv$ should align with $\xv$. 
Therefore, we can straightforwardly devise the consistency metric based on the information alignment defined above:
\vspace{-15pt}
\begin{equation}
\small
\vspace{-5pt}
\begin{split}
    \textsc{Consistency}(\yv, \xv) = \mathrm{mean}\left( \textit{align}(\yv \to \xv) \right), 
\end{split}
\end{equation}
which is the average alignment scores of tokens in $\yv$ w.r.t. $\xv$. 
Our metric offers a simpler solution than the recent QA-based metrics \cite{scialom-etal-2019-answers,Durmus_2020,wang2020asking} that compare 
the answers extracted from $\yv$ and $\xv$ by a Question-Answering system, and is more interpretable than the black-box consistency classification models \cite{falke-etal-2019-ranking,kryscinski2019evaluating,maynez2020faithfulness}. We also achieve stronger empirical performance (\S\ref{sec:compression-experiments}).

\paragraph{Relevance}
As one of the most heavily studied aspects of summarization, relevance concerns how well the summary $\yv$ retains important information in $\xv$ \cite{nenkova-passonneau-2004-evaluating,zopf-etal-2016-beyond}. As in previous work, the ``importance'' of information can be determined by human-written reference summaries $\rv$. That is, a piece of information is considered important if it is mentioned in a reference. The intuition can readily be captured by the information alignment $\textit{align}(\rv\to\yv)$ that measures the extent to which information in reference $\rv$ is covered by the summary $\yv$. 
Additionally, we account for the criterion that any information in $\yv$ should be precise, i.e., consistent with $\xv$. Combining the two considerations, the full definition of our relevance metric conveys the intuition that a fully relevant summary $\yv$ should \textit{achieve both and balance} reference-alignment and consistency:
%
%
\begin{equation}
\small
\begin{split}
    &\textsc{Relevance}(\yv, \xv, \rv) = \\
    &\mathrm{mean}\left( \textit{align}(\rv\to\yv) \right) \times \mathrm{mean}\left( \textit{align}(\yv\to\xv) \right),
\end{split}
\label{eq:metric:relevance}
\end{equation}
which is the product of both components. 
Traditional reference-based metrics consider only the reference text (rather than the input). For example, ROUGE~\cite{lin-2004-rouge} can be seen as measuring the alignment between $\yv$ and $\rv$ where the alignment is defined by text matching. Our metric, with the combination of both reference and input, plus better alignment modeling (\S\ref{sec:alignment_implementations}), greatly outperforms those previous metrics (\S\ref{sec:compression-experiments}).

\subsection{Evaluation of ``Transduction'' Tasks} \label{sec:transduction}
We take style transfer as the example task to discuss semantic preservation of transduction tasks.
The aim of style transfer is to generate text $\yv$ that changes one or more stylistic attributes (e.g., formality) of source text $\xv$ and completely preserve its style-independent information \cite{hu2017toward,NIPS2017_yelp}. Measuring content preservation is the core yet challenging problem for the evaluation.


\paragraph{Preservation} 
A transduction result $\yv$ is required to contain \emph{all and only} information from  $\xv$. 
In other words, all tokens in $\yv$ should align with $\xv$, and \emph{vice versa}. Considering the former to be the ``precision'' of the $\yv$ information w.r.t $\xv$, and the latter the ``recall'', we naturally arrive at the following ``F1''-style definition of the preservation metric:
\vspace{-2pt}
\begin{equation}
\small
\begin{split}
    &\textsc{Preservation}(\yv, \xv) = \\
    &\frac{\mathrm{mean}\left( \textit{align}(\yv\to\xv) \right) \times \mathrm{mean}\left( \textit{align}(\xv\to\yv) \right)}{\mathrm{mean}\left( \textit{align}(\yv\to\xv) \right) + \mathrm{mean}\left( \textit{align}(\xv\to\yv) \right)}, 
\end{split}
\label{eq:metric:preservation}
\end{equation}
which is the harmonic mean of the two directions of information alignment.
Note that the two-way alignments differ from the ``consistency'' and ``relevance'' metrics in compression where we have only required output $\yv$ to align with input $\xv$. Our experiments show that it is crucial to account for alignments in both directions for transduction (\S\ref{sec:transduction-experiments}). 

\begin{figure*}[t]
    \centering
    \includegraphics[width=\textwidth]{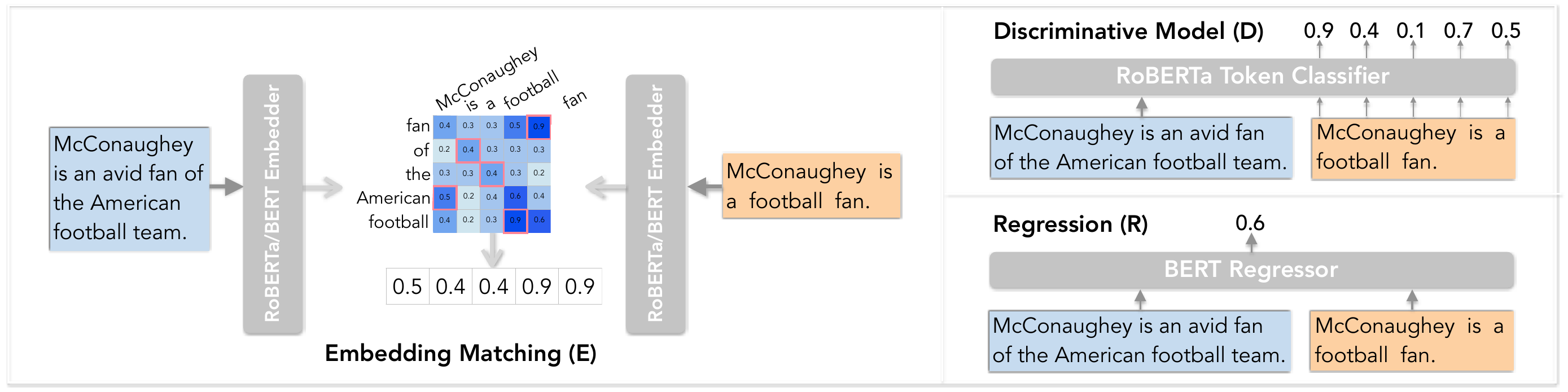}
    \vspace{-24pt}
    \caption{\small We study three effective ways of information alignment prediction, i.e., embedding matching (left), discriminative model (upper right) and regression (lower right). The figure illustrates the estimation of alignment from {\color{orange} output} to {\color{blue} input}.
    }
    \label{fig:alignment_prediction_models}
\end{figure*}

\subsection{Evaluation of ``Creation'' Tasks} \label{sec:creation}
We formulate aspects of creation tasks using the example of  knowledge-grounded dialog generation. In this task, an agent generates text $\yv$ as a response to conversation history $\xv$ while exhibiting information from knowledge context $\cv$, e.g., an external document~\citep{qin2019conversing,guo2018topicbased} or a set of facts~\citep{dinan2018wizard,zhang-etal-2018-personalizing}.
For the agent, sustaining an engaging conversation is considered an essential skill~\cite{venkatesh2018evaluating,guo2018topicbased,mehri2020usr}. Besides, the generated response must be grounded in the knowledge context by referring to its information as often as possible~\cite{dinan2019second,smith-etal-2020-put}. 
We devise metrics for the two central aspects, respectively.

A crucial property of creation tasks is that the agent is allowed to create new information beyond the input and context. Thus, to aggregate the information alignment vector, it is more suitable to consider the \emph{total volume} rather than the density. That is, we would use $\mathrm{sum}(\cdot)$ instead of the previous $\mathrm{mean}(\cdot)$ to aggregate token-level alignment scores. 


\paragraph{Engagingness} 
We adopt the common definition of engagingness \cite[e.g.,][]{mehri2020usr}, namely,  the response should not be generic or dull (e.g., \emph{``I don't know''}), but engages the partner in conversation, such as presenting an interesting fact. 
Therefore, an engaging response $\yv$ should provide high volume of information that acknowledges both the history $\xv$ to engage the partner and the context $\cv$ which we assume contains relevant facts. This naturally leads to the following metric definition:
\begin{equation}
\small
\begin{split}
    \textsc{Engagingness}(\yv, \xv, \cv) =
    \mathrm{sum} \left( \textit{align}( \yv \rightarrow [\xv,\cv] ) \right) , 
\end{split}
\label{eq:metric:engaging}
\end{equation}
where we concatenate the history $\xv$ and knowledge context $\cv$, and measure the extent of response $\yv$'s acknowledgement of the information.
Previous works have devised various metrics for the aspect, ranging from 
measuring response-topic consistency~\cite{guo2018topicbased}, conversation length~\cite{venkatesh2018evaluating}, retrieval of reference responses~\cite{mehri2020usr}, etc. Our metric is cleanly defined in line with all other metrics we developed, and shows stronger human correlation than previous designs.

\paragraph{Groundedness} 
As a widely studied aspect of knowledge-based dialog, groundedness measures how well the response refers to the knowledge context \cite{dinan2018wizard,qin2019conversing,mehri2020usr}.
Straightforwardly, the aspect can be evaluated with the following metric:
\vspace{-15pt}
\begin{equation}
\small
\begin{split}
    \textsc{Groundedness}(\yv, \cv) = \mathrm{sum}\left( \textit{align}(\yv \rightarrow \cv)\right), 
\end{split}
\end{equation}
which measures the alignment between the response $\yv$ and knowledge context $\cv$.

\subsection{Implementation of Alignment Estimation} \label{sec:alignment_implementations}
We have presented the metrics for a range of key aspects in different tasks, building on the core information alignment measure (Definition~\ref{def:info-align}). We next discuss different effective implementations for measuring the alignment scores between text, including embedding matching, discriminative model, and regression, all based on powerful pretrained language models (Figure~\ref{fig:alignment_prediction_models}). 
%



\paragraph{Embedding Matching (\texttt{E})}
One simple way to estimate the alignment vector $\textit{align}(\av\to\bv)$ is by matching the embeddings of tokens in the two sequences. Specifically, we use either pretrained BERT \cite{devlin-etal-2019-bert} or RoBERTa \cite{liu2019roberta} to extract contextual embedding for each token in $\av$ and $\bv$, normalize each embedding vector to unit norm, and then use greedy matching following \cite{corley-mihalcea-2005-measuring,zhang2020bertscore}. That is, the alignment score of each token in $\av$ is defined as its maximum cosine similarity with the tokens in $\bv$. 
We found in our empirical studies  (\S\ref{sec:exp}) that the \texttt{E} method seems to work better when $\av$ and $\bv$ have similar volume of information (so that one-to-one token matching is suitable).

\begin{figure*}[t]
\centering
\includegraphics[page=1,width=0.43\textwidth]{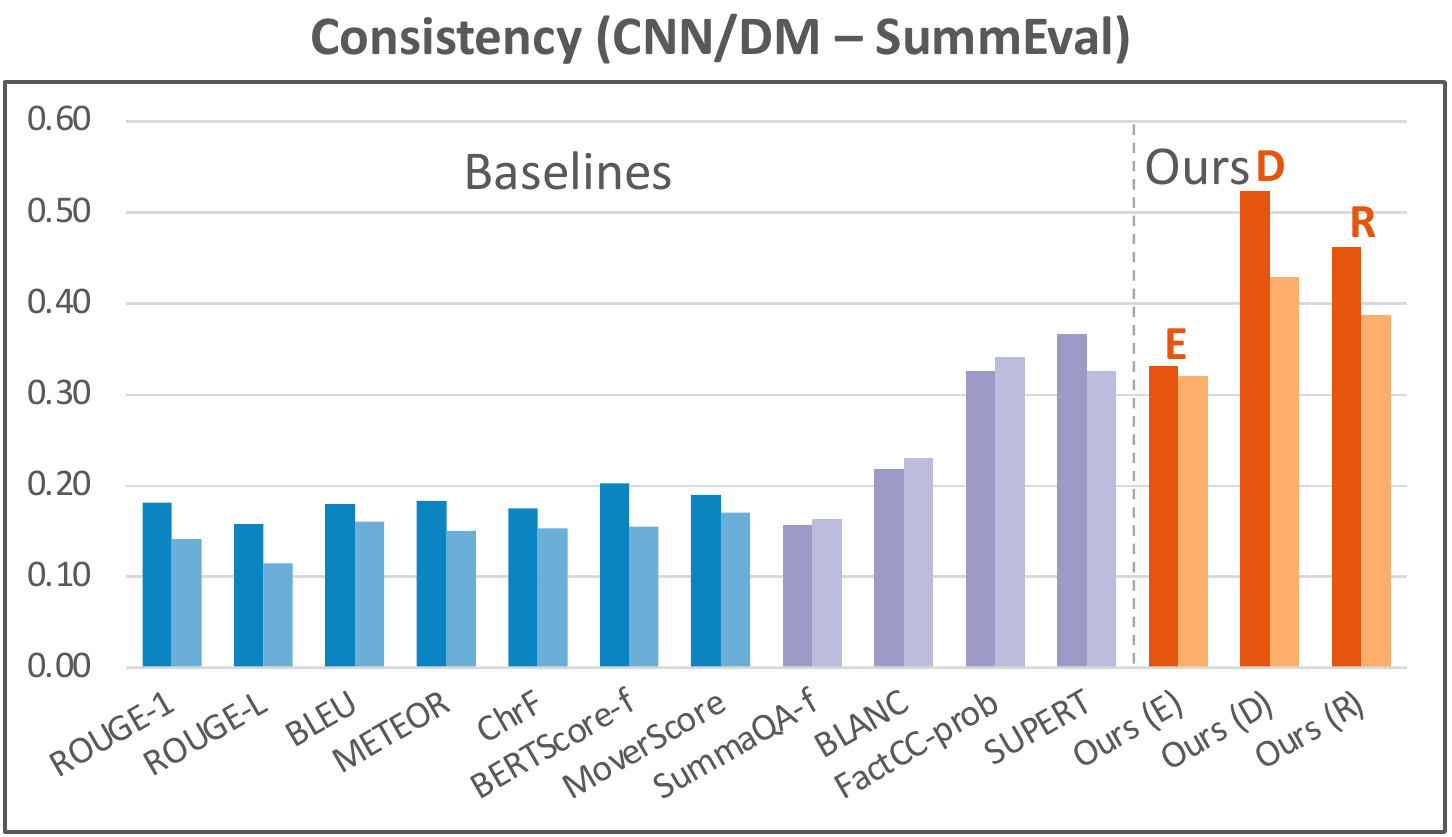}
\hspace{0.07\textwidth}
\includegraphics[page=2,width=0.43\textwidth]{figs/compression_graph_camera_ready.pdf}
\vspace{-10pt}
\caption{\small Correlations with human judgement on \emph{consistency} in summarization. \texttt{E} denotes our metrics using embedding-matching alignment estimation, \texttt{D} using discriminative-model and \texttt{R} using regression. Reference-based metrics are in {\color{blue} blue}, reference-free metrics in {\color{Orchid} purple}, and our metrics in {\color{red} red}/{\color{orange} orange}. For SummEval annotation data \cite{fabbri2021summeval} {\bf (left)}, we report Pearson (left, dark color) and Spearman (right, light color) correlations for each metric.
For QAGS annotation data \cite{wang2020asking} {\bf (right)}, only Pearson correlations were available for baselines.
}
\label{fig:consistency_graph}
\end{figure*}

\paragraph{Discriminative Model (\texttt{D})}
To estimate the information alignment from arbitrary text $\av$ to $\bv$, we formulate the problem as sequence tagging, for which we train a model that labels each token in $\av$ with $1$ if it aligns with $\bv$, and $0$ otherwise. The predicted probability of label $1$ for each $\av$ token serves as the alignment score. We base our model on RoBERTa and train with automatically constructed weak supervision data. Appendix \S\ref{sec:appendix} describes all details. For example, to learn to estimate the alignment of the output $\yv$ to the input in an NLG task, we use the training corpus of the task: 
for each output $\yv$, we perturb it by masking out portions of tokens and using a pretrained BART \cite{lewis-etal-2020-bart} model to fill in the blanks~\cite{zhou2020detecting}. The BART model is not conditioning on any input context (e.g., $\xv$), so the infilled tokens can be considered to not align with the input. We do the masking by first applying constituency parsing to the text and then randomly masking out a subtree of the parsing. Besides the infilling data, we also augment the training with paraphrasing data. That is, we apply a paraphrasing model to $\yv$, and treat all tokens in the paraphrases as alignment to the input. 
Note that $\yv$ need not be the gold output, but can also be any automatically constructed output as long as it is guaranteed to align fully with the input. For example, an output $\yv$ by an extractive summarization model aligns fully with the input article. We will see more examples in our experiments.

\begin{figure}[t]
\centering
\includegraphics[page=3,width=0.43\textwidth]{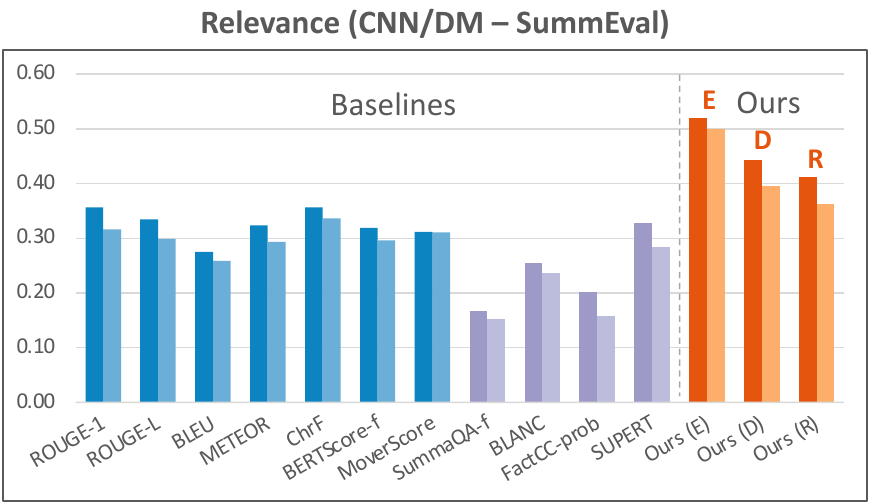}
\vspace{-10pt}
\caption{ \small
Correlations with human judgments on relevance in summarization. Reference-based metrics are computed using all 11 references for each example provided in the data. The plot format follows Figure~\ref{fig:consistency_graph}.
}
\label{fig:relevance_graph}
\end{figure}

\begin{table}[t]
\vspace{5pt}
\small
\centering
\begin{tabular}{lllll}
\toprule
{Relevance} &
  \multicolumn{1}{l}{$\rv\to\yv$} &
  \multicolumn{1}{l}{$\yv\to\xv$} &
  \multicolumn{1}{c}{$+$} &
  \multicolumn{1}{l}{$\times$ (Ours)} \\ \hline
Align (E) & 0.4705 & 0.4381 & 0.5195 & \textbf{0.5198} \\
Align (D) & 0.4184 & 0.2834 & 0.4308 & \textbf{0.4423} \\
Align (R) & 0.4050 & 0.2688 & 0.3861 & \textbf{0.4115} \\ \bottomrule
\end{tabular}
\vspace{-10pt}
\caption{\small
Ablation Studies: Pearson correlations for different variants of our \emph{relevance} metric (Eq.\ref{eq:metric:relevance}) using different components and combination strategies. $\rv\to\yv$ corresponds to $\mathrm{mean}\left( \textit{align}(\rv\to\yv) \right)$ and similarly for $\yv\to\xv$; $+$ sums the two components and $\times$ is our design that takes the product. 
} 
\label{table:relevance-combination}
\end{table}

\paragraph{Aggregated Regression (\texttt{R})} 
Instead of estimating the per-token alignment vector as defined in Eq.\eqref{eq:alignment-def}, we may also directly estimate the single aggregated alignment score such as $\mathrm{mean} \left( \textit{align}( \av \rightarrow \bv ) \right)$ (or $\mathrm{sum}$). This is because all the metrics proposed above have only used the aggregated score. To this end, we train a regression model using the same weak supervision data for \texttt{D}, with the aggregated alignment score as the regression target. Similar to \citet{sellam-etal-2020-bleurt}, in our experiments, we implement the regression model with BERT \cite{devlin-etal-2019-bert}. In particular, we initialize the regression model with the intermediate \texttt{BERT-base-midtrained} model weights provided by \citet{sellam-etal-2020-bleurt}. 
We note that the aggregated estimation method may not be applicable to future metrics in our evaluation framework when fine-grained per-token alignment is required.

\section{Experiments}\label{sec:exp}

We evaluate the proposed metrics on commonly used human annotation datasets for summarization (\S\ref{sec:compression-experiments}), style transfer (\S\ref{sec:transduction-experiments}) and dialog (\S\ref{sec:creation-experiments}), and study the effect of information alignment accuracy on the performance of metrics (\S\ref{sec:ablation-experiments}).

\paragraph{Evaluation Criteria}
To measure a metric's performance on an aspect, we compute the sample-level correlation between the metric scores and human judgments on generation samples.
We also evaluate system-level correlation (based on the ranking of comparison systems) as the secondary criterion \cite{mathur2020tangled} and report results in the appendix, which typically exhibits the same patterns as sample-level correlation.
We measure Pearson and Spearman correlations whenever applicable. 
We also report Kendall-Tau correlation in the appendix when available. 



\subsection{Experiments for ``Compression'' Metrics} \label{sec:compression-experiments}

\paragraph{Datasets} 
For the \emph{consistency} aspect, we follow previous studies and evaluate metrics using human annotations from two commonly-used sources: {\bf (1)} SummEval \cite{fabbri2021summeval} on the CNN/DM summarization dataset \cite{hermann2015teaching,nallapati-etal-2016-abstractive}. The annotation dataset contains 1,600 examples from 16 summarization systems;
{\bf (2)} QAGS \cite{wang2020asking} (which names the aspect ``correctness'') on the XSUM dataset \cite{narayan-etal-2018-dont}, another summarization task with strong abstractive property. The dataset contains 235 outputs from a fine-tuned BART model \cite{lewis-etal-2020-bart}. The QAGS dataset also contains another 239 outputs for CNN/DM, for which we report results in Table \ref{table:qags-consistency} in the appendix. 

For \emph{relevance}, we test our metric on the respective annotations from SummEval on CNN/DM. 


\paragraph{Baselines and Setup} 
For baselines, we include commonly-used metrics reported in previous papers, ranging from reference-based metric, such as ROUGE, BLEU and BERTScore \cite{zhang2020bertscore}, to reference-free ones, such as SummaQA \cite{scialom-etal-2019-answers} based on QA and FactCC \cite{kryscinski2019evaluating} based on sentence classification. 
For our metrics, we use \texttt{RoBERTa-large} for the embedding-matching (\texttt{E}) alignment since it was pre-trained on the CommonCrawl News dataset \cite{nagel2016ccnews} that is close to the summarization domains. For the discriminative-model (\texttt{D}) alignment, we train two \texttt{RoBERTa-large} token classifiers to compute $\textit{align}(\yv\to\xv)$ and $\textit{align}(\rv\to\yv)$, respectively, with training data automatically constructed for CNN/DM and XSUM according to Appendix \S\ref{subsec:appendix-compression}. 
For the regressive (\texttt{R}) alignment, we train the \texttt{BERT} models (\S\ref{sec:alignment_implementations}) to estimate the respective mean alignment scores. 

\paragraph{Results}
We present the consistency results in Figure \ref{fig:consistency_graph}. On CNN/DM, our metrics based on the trained alignment models (\texttt{D} and \texttt{R}) both clearly outperform previous metrics.  
On XSUM, our \texttt{D}-based metric 
also achieves the best performance. 
The 
\texttt{E}-based metric sees a catastrophic drop in correlations, which is likely due to the higher abstractiveness of XSUM summaries that renders embedding matching inadequate.
The sentence-classifier based \texttt{FactCC} metric \cite{kryscinski2019evaluating}, which is trained to distinguish paraphrases from artificially perturbed sentences, also achieves a decent correlation on XSUM. However, it seems unable to effectively model the summaries on CNN/DM that tend to be longer and richer in information, and thus produces a lower correlation.


Figure~\ref{fig:relevance_graph} shows the results for relevance on CNN/DM. Our metrics strongly outperform all other baselines, showing that accounting for alignments with both references and the input article (Eq.\ref{eq:metric:relevance}) is superior to only considering the references (metrics in blue in the figure) or the input article (metrics in purple). This is further validated by the ablation studies in Table \ref{table:relevance-combination}, which demonstrate that multiplying the two alignments, which emphasizes joint and balanced achievement of both, improves the correlations compared to individual alignments or simply summing them together.
Figure~\ref{fig:relevance_graph} also shows our 
\texttt{E}-based implementation performs better than the 
\texttt{D}- and \texttt{R}-based variants, likely because the metric involves alignment between generation and references which tend to have similar information volume and thus favor one-to-one token mapping. We observe similar patterns in transduction below.


\begin{figure}[t]
\centering
\includegraphics[width=0.44\textwidth]{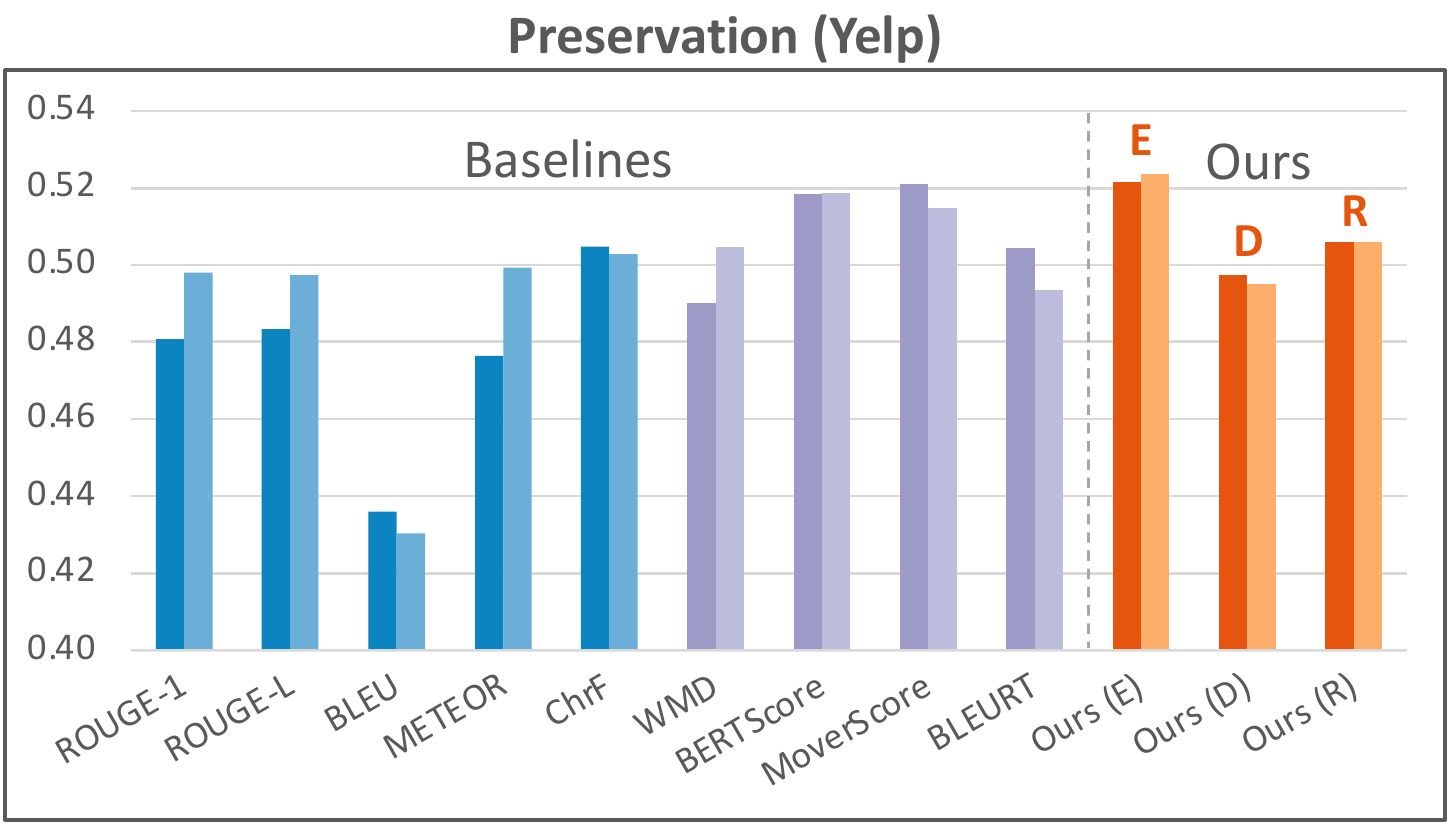}
\vspace{-10pt}
\caption{\small
Human correlations on style transfer \emph{preservation} aspect.
Lexical-matching metrics are in {\color{blue} blue}. Embedding- or model-based similarity metrics are in {\color{Orchid} purple}, and ours are in {\color{red} red}/{\color{orange} orange}. 
}
\label{fig:transduction_graph}
\end{figure}

\begin{table}[t]
\vspace{5pt}
\small
\centering
\begin{tabular}{llll}
\toprule
 {Preservation}
   & \multicolumn{1}{l}{$\yv\to\xv$} & \multicolumn{1}{l}{$\xv\to\yv$} & \multicolumn{1}{l}{$\yv\Leftrightarrow\xv$ (Ours)} \\ \hline
Align (E) & 0.4989 & 0.5078 & {\bf 0.5216} \\
Align (D) & 0.4481 & 0.4608 & {\bf 0.4974} \\
Align (R) & 0.4744 & 0.4823 & {\bf 0.5060} \\
\bottomrule
\end{tabular}
\vspace{-10pt}
\caption{ \small
Ablation Studies: Pearson correlations for variants of \emph{preservation} metric (Eq.\ref{eq:metric:preservation}) accounting for different directions of information alignment. $\yv\to\xv$ corresponds to $\mathrm{mean}\left( \textit{align}(\yv\to\xv) \right)$ and similarly for $\xv\to\yv$; Our $\yv\Leftrightarrow\xv$ is harmonic mean of alignments in both directions.
}
\label{table:preservation-f1}
\end{table}

\begin{figure*}[t]
\centering
\includegraphics[page=1,width=0.43\textwidth]{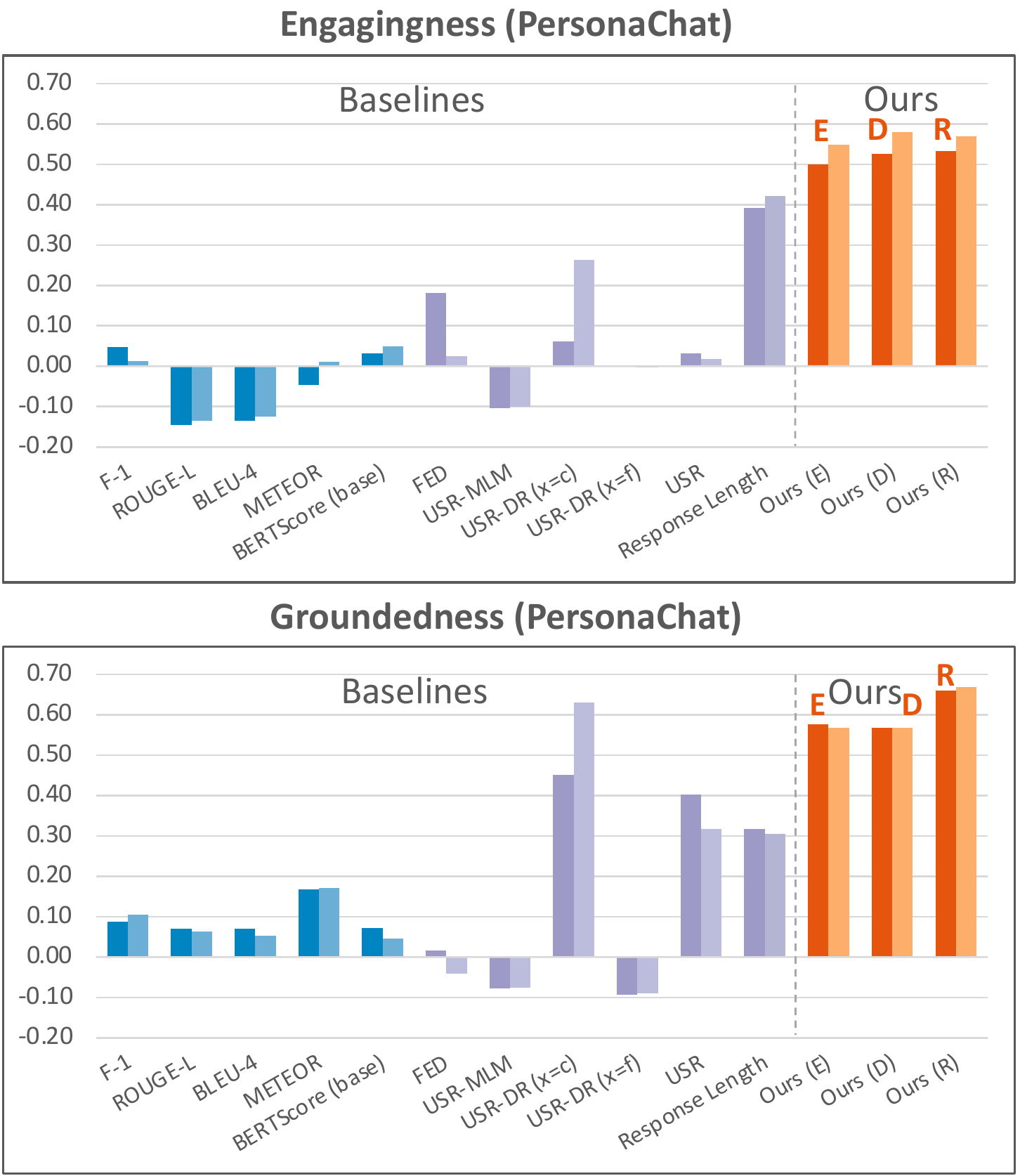}
\hspace{0.07\textwidth}
\includegraphics[page=2,width=0.43\textwidth]{figs/dialog_graphs_camera_ready.pdf}
\vspace{-10pt}
\caption{Correlations with human judgement on \emph{engagingness} and \emph{groundedness} aspects for knowledge-grounded dialog. 
The plot format is the same as Figure~\ref{fig:consistency_graph}.
}
\label{fig:creation_graph}
\end{figure*}

\subsection{Experiments for ``Transduction'' Metrics}
\label{sec:transduction-experiments}

\paragraph{Datasets} 
We apply our \emph{preservation} metric to evaluating text style transfer, where we use the human annotations from \cite{mir-etal-2019-evaluating} on the Yelp sentiment transfer data \cite{NIPS2017_yelp}\footnote{It is arguable whether ``sentiment'' is part of style \cite{krishna2020reformulating}. Here we just use the most common dataset.}.
The dataset contains 8,784 outputs from 12 systems. 

\paragraph{Baselines and Setup} 
We compare with the common metrics used previously \cite{mir-etal-2019-evaluating}, and further include BERTScore \cite{zhang2020bertscore}, MoverScore \cite{zhao2019moverscore} and BLEURT \cite{sellam-etal-2020-bleurt}, the latest neural text similarity metrics. We use BLEURT out-of-the-box as we do not assume access to human scores for finetuning the evaluation models. 
For our metrics, we use \texttt{RoBERTa-large-MNLI} for embedding-matching (\texttt{E}) due to its fine-tuning on entailment detection which is close to the domain under study.
For discriminative model (\texttt{D}), we train \texttt{RoBERTa-large} on Yelp alignment data created by paraphrasing and perturbing the inputs $\xv$.
For regression (\texttt{R}), we train to estimate the mean alignment score computed from the same dataset as \texttt{D}. 

\paragraph{Results}
We present preservation results in Figure~\ref{fig:transduction_graph}. Our metric (\texttt{E}) achieves competitive or better performance than all previous metrics.
MoverScore \cite{zhao2019moverscore} as a strong baseline computes word mover's distance \cite{kusner2015word} between input $\xv$ and output $\yv$ token embeddings. In contrast, our metric explicitly accounts for the two-way input-output alignments with an ``F1''-style harmonic mean aggregation (Eq.\ref{eq:metric:preservation}). Table~\ref{table:preservation-f1} shows the two-way approach is effective and exhibits higher correlation compared to single-directional alignment, in line with the nature of transduction tasks.
Similar to their relevance results in summarization, our 
\texttt{D}- and \texttt{R}-based implementations fall behind \texttt{E}, likely because token matching is more suitable for measuring alignments between two text pieces with similar information volume.


\begin{table}[t]
\small
\centering
\begin{tabular}{lllll}
\toprule
Engagingness & \multicolumn{2}{l}{Mean} & \multicolumn{2}{l}{Sum (Ours)} \\
             & P       & T        & P       & T         \\ \hline
Align (E)     & 0.1502     & 0.3184    &  \textbf{0.5003}     & \textbf{0.4937}    \\
Align (D)     & 0.1821     & 0.3223    &  \textbf{0.5265}    & \textbf{0.5163}    \\
Align (R)     & -0.0490     &  -0.0191   &  \textbf{0.5320}   & \textbf{0.4653}    \\ \bottomrule
\end{tabular}
\vspace{-10pt}
\caption{
\small
Ablation Studies: Pearson correlations for our engagingness metric (Eq.\ref{eq:metric:engaging}) with different alignment aggregation strategies. ``Mean'' takes the average of the alignment vector, and ``Sum'' is our designed metric that takes the sum. ``P'' and ``T'' refer to PersonaChat and TopicalChat, respectively.
}
\label{table:dialog-aggregation}
\end{table}


\subsection{Experiments for ``Creation'' Metrics}
\label{sec:creation-experiments}

\paragraph{Datasets} 
For the \emph{engagingness} aspect, we use the latest human annotation data collected by \cite{mehri2020usr} (which names the aspect ``interesting'')
on PersonaChat \cite{zhang-etal-2018-personalizing} and TopicalChat \cite{Gopalakrishnan2019}, two knowledge-grounded dialog tasks with different forms of knowledge. The dataset contains 300 examples from 5 systems for PersonaChat, and 360 examples from 6 systems for TopicalChat. All turns preceding the current response $\yv$ are treated as the history $\xv$ ($4.2$ turns on average for PersonaChat and $5.1$ turns for TopicalChat). The knowledge context $\cv$ refers to the persona statements in PersonaChat and the knowledge snippets in TopicalChat.

For the \emph{groundedness} aspect, we again use the human annotations from \citet{mehri2020usr} (which names the aspect ``uses knowledge'') on both PersonaChat and TopicalChat. 


\paragraph{Baselines and Setup}
We compare with all the diverse metrics studied in \cite{mehri2020usr} and FED \cite{mehri2020unsupervised}, a set of latest unsupervised dialogue metrics based on the DialoGPT model \cite{zhang2019dialogpt}.
We use FED-Interesting from the original paper designed for engagingness and FED-Informative designed for groundedness, respectively. We also add a particularly simple baseline---response length, which as we show performs surprisingly well. 
For our metrics, we use \texttt{BERT-base} for embedding matching (\texttt{E}), \texttt{RoBERTa-large} token classifiers trained on $\textit{align}(\yv\to[\xv,\cv])$ and $\textit{align}(\yv\to\cv)$ for discriminative model (\texttt{D}), and \texttt{BERT-base} regressors on the sums of the respective alignment scores
for regression (\texttt{R}). We create separate alignment datasets for PersonaChat and TopicalChat, as described in Appendix \ref{subsec:appendix-creation}. 

\begin{figure}[t]
    \centering
    \includegraphics[width=0.5\textwidth]{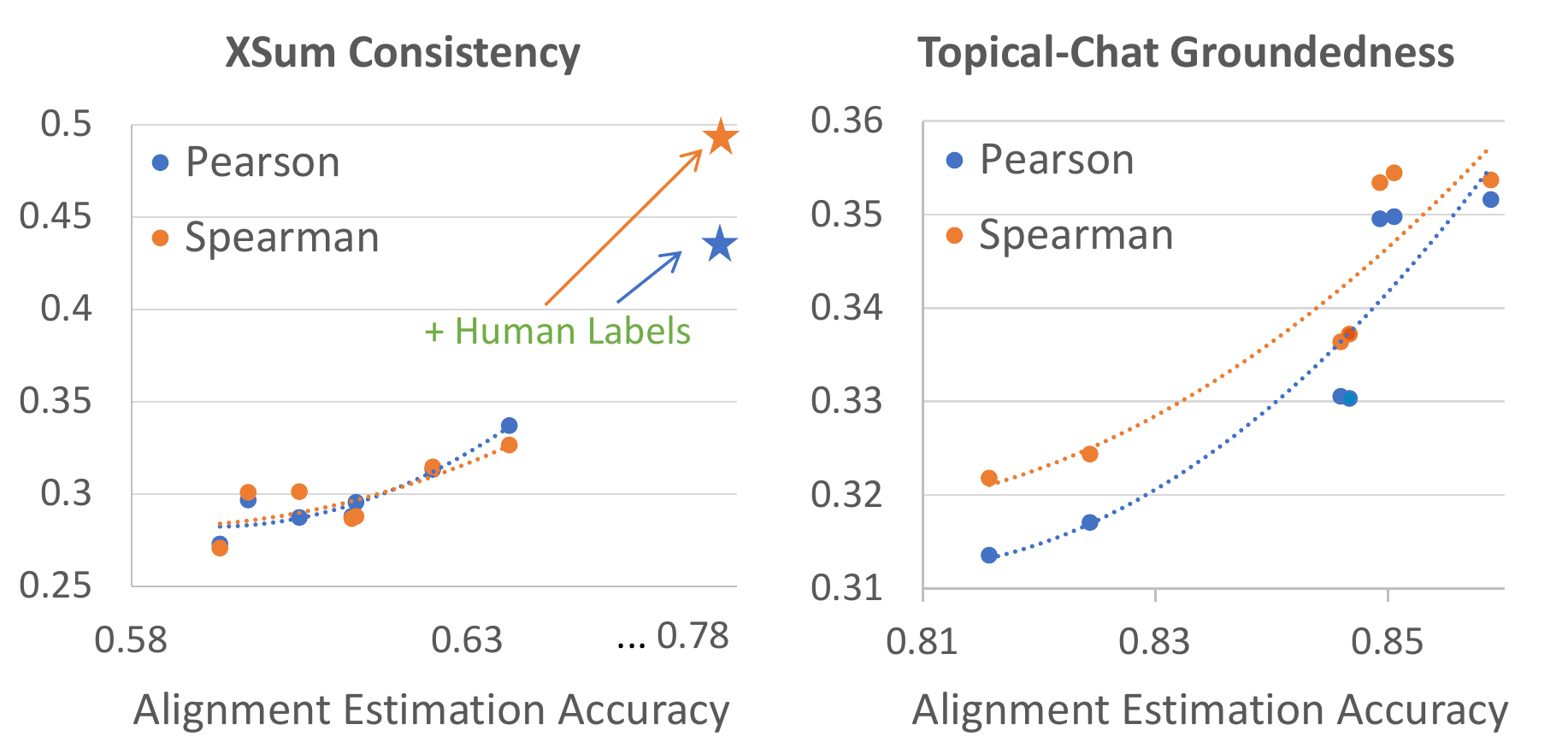}
    \vspace{-20pt}
    \caption{Effect of alignment estimation accuracy on metric performance.}
    \label{fig:hal_corr}
\end{figure}

\paragraph{Results}
We present the results for engagingness in the top two plots of Figure~\ref{fig:creation_graph}. Our metrics with different implementations all improve over previous methods by large margins on the two datasets.
Many of the baseline metrics show decent correlations on TopicalChat, but fail on the PersonaChat corpus. This is likely because PersonaChat requires strong dependency of responses on the dialog history and knowledge context, thus metrics that do not directly model the dependency (e.g., USR-DR \cite{mehri2020usr} based on response retrieval) as ours struggle for accurate evaluation. 

Noticeably, the simple \texttt{response length} performs consistently well on both datasets, far better than previous metrics on PersonaChat. The baseline can be considered as a special case of ours by setting alignment scores of all tokens to $1$. The stronger correlations of our model-based metrics demonstrate the effect of accurate alignment. 

Ablation studies in Table \ref{table:dialog-aggregation} shows that measuring the volume ($\mathrm{sum}$) instead of the density ($\mathrm{mean}$) of aligned information is crucial for the superior performance of our metrics, highlighting the unique characteristics of the ``creation'' task (\S\ref{sec:creation}). 


The results for groundedness are shown in the bottom two plots of Figure~\ref{fig:creation_graph}. Our metrics again generally achieve strong correlations, with the \texttt{R}-based metric consistently outperforming other implementations, likely because the estimation of grounded information \textit{volume} ($\mathrm{sum}$) benefits from the expressivity of end-to-end models. This is indicated by the underperformance of the \texttt{D}-based metric, which is trained on the same data but aggregates token-level predictions with more structure.


We provide more empirical studies in Appendix \S\ref{sec:appendix:dialog}. In particular, we found that besides the two core aspects, our alignment based method also achieves stronger human correlations than existing metrics on other dialog aspects, such as the \emph{understandability} and \emph{naturalness} of responses (Table~\ref{table:additional_correlations}). 


\subsection{Ablation: higher alignment estimation accuracy, better correlation}
\label{sec:ablation-experiments}

We study how the accuracy of information alignment estimation influences the performance of metrics. 
We demonstrate a highly  desirable pattern that higher alignment estimation accuracy can usually lead to better correlation. This indicates that improvement on the single alignment estimation model could immediately benefit a broad range of aspect metrics defined in our unified framework.

Specifically, we use the discriminative model (\S\ref{sec:alignment_implementations}) for our study. First, we vary the number of training iterations to get different model checkpoints, and evaluate both the alignment estimation accuracy and the metric human correlation based on the checkpoints. We evaluate accuracy with the human-annotated token alignment labels on the XSUM summarization data \citet{maynez2020faithfulness}. Figure~\ref{fig:hal_corr} (left) shows the \emph{consistency} metric achieves better correlation as the alignment accuracy increases. We do the same on TopicalChat dialog data and evaluate accuracy with our weak supervision data (since no human labels are available). Figure~\ref{fig:hal_corr} (right) shows similar trends for the \emph{groundedness} metric. Second, we further use part of XSUM human alignment annotations to finetune the alignment model, and obtain even higher accuracy, which in turns gives better correlation for \emph{consistency} evaluation (star marks in the figure).


\section{Conclusions}
We have proposed a general evaluation framework for NLG tasks categorized as compression, transduction, and creation. Based on the common central concept of information alignment between input, context, and output, we devised a family of interpretable metrics for the key aspects of diverse tasks (summarization, style transfer, and dialog) \mingkai{uniformly in terms of the alignment, most of which don't require human references}. The uniformly designed metrics achieve superior or comparable human correlations compared to existing metrics. 

The unified framework not only offers structured guidance for the metric design of new aspects and tasks, \mingkai{but also opens up exciting possibilities for composable NLG evaluation. Following the more rigorous practices of software engineering, we may divide the process into modular components that can be improved, scaled, and diagnosed independently.} We are excited to explore more in the future.

\section*{Acknowledgements}
Bowen Tan is sponsored by US NGA NURI No. HM0476-20-1-0002 and the National Science Foundation under Grant No. IIS-16-17583, IIS-19-55532 and CNS-20-08248. Any opinions, findings, and conclusions or recommendations expressed in this material are those of the authors and do not necessarily reflect the views of the NGA or the U.S. Government.

\bibliography{anthology,custom}
\bibliographystyle{acl_natbib}

\clearpage 

\appendix


\setcounter{page}{1}

\numberwithin{equation}{section}
\numberwithin{table}{section}
\numberwithin{figure}{section}
\setlength{\belowcaptionskip}{0pt} 

\section{Implementation of Alignment Estimation Models}
\label{sec:appendix}


\mingkai{We train our alignment models by constructing} weakly supervised data using texts in the domain of evaluation. The data construction process can be divided into three steps:
\begin{enumerate}
    \item Retrieve or generate a target sentence $\yv_1$ given the desired input $\xv$ (e.g., the document in summarization tasks). All tokens in $\yv_1$ should be considered aligned with $\xv$;
    \item Sometimes, $\yv_1$ consists of several original sentences from $\xv$. In order to make our model non-trivial and more robust, we generate a pharaphrase $\yv_2$ of $\yv_1$ \mingkai{with a pretrained paraphrase generator\footnote{\url{https://huggingface.co/Vamsi/T5\_Paraphrase\_Paws}};}
    \item After that, we mask some portion of $\yv_2$, and use a \texttt{BART-large} model \cite{lewis-etal-2020-bart} to infill those masks. Because the infilled content is generated without conditioning on $\xv$, we label the infilled words as "not aligned" with $\xv$ (BAD), and other words of $\yv_2$ are labeled as "aligned" (OK);
\end{enumerate}
Finally, $\xv$, $\yv_2$, and alignment labels on $\yv_2$'s words are our desired \mingkai{training} data.

Specially on our paraphrasing operation, in order to make the generated paraphrase different enough from the original text, we always generate 10 paraphrases and take the one with biggest edit distance with the sentence; and specially about our masking mechanism, we randomly mask some sub-trees in the constituency parsing tree of $\yv_2$ \mingkai{with a pretrained parser \footnote{\url{https://github.com/nikitakit/self-attentive-parser}}.}
The differences across tasks are \mingkai{the definitions of $\xv$ and $\yv_1$} in the step (1), as detailed below.

\subsection{Compression: Summarization}
\label{subsec:appendix-compression}
Our training for $\textit{align}(\yv\to\xv)$ in the summarization domain is reference-free. We use the document as $\xv$, \mingkai{and generate its pseudo-summaries as $\yv_1$} using a traditional unsupervised extractive summarizer based on TextRank \cite{mihalcea-tarau-2004-textrank}. \mingkai{We don't use reference summaries because they can contain hallucinations that don't align with the article \cite{maynez2020faithfulness}. In an ablation study with XSUM Consistency data~\cite{wang2020asking}, training a \texttt{D} model using reference summaries leads to $0.2822$ Pearson correlation compared to $0.3222$ using auto-generated summaries, which is clearly lower.} To train for $\textit{align}(\rv\to\yv)$, we use the reference as both $\xv$ and $\yv_1$.

\subsection{Transduction: Text Style Transfer}
\label{subsec:appendix-transduction}
In this domain, we simply set $\yv_1$ to be the original sentence $\xv$.

\subsection{Creation: Dialog}
\label{subsec:appendix-creation}
When training for $\textit{align}(\yv\to[\xv,\cv])$, we use the reference response as $\yv_1$ and the  concatenation of $\xv$ and $\cv$ as the input. \mingkai{For models that predict $\textit{align}(\yv\to\cv)$, we set the knowledge context $\cv$ as the input, and randomly extract sentences from it as $\yv_1$. For PersonaChat, we sample 1-3 sentences at random, whereas for TopicalChat, we only sample 1 sentence because its $\cv$ tends to be long. When aggregating the alignment vectors, we remove stopwords according to NLTK \cite{bird2009natural} to focus on important words.}

\clearpage

\onecolumn
\section{Key Aspects}
\begin{table*}[h]
\centering
\begin{tabular}{|l|l|l|l|}
\hline
\textbf{Task Category}       & \textbf{Aspect} & \textbf{Alternative Names}                                                                             & \textbf{Considered By} \\ \hline
\multirow{2}{*}{Compression} & Consistency     
& \begin{tabular}[c]{@{}l@{}}Factual Correctness, \\ Faithfulness, \\ (No) Hallucination \end{tabular} 
&  \begin{tabular}[c]{@{}l@{}} \cite{wang2020asking}, \\ \cite{maynez2020faithfulness}, \\ \cite{Durmus_2020}, \\ \cite{kryscinski2019evaluating}, \\ \cite{fabbri2021summeval}, etc. \end{tabular}                \\ \cline{2-4} 
                             & Relevance       & \begin{tabular}[c]{@{}l@{}} Content Selection, \\ Importance \end{tabular}                       
&  \begin{tabular}[c]{@{}l@{}} \cite{nenkova-passonneau-2004-evaluating} \\ \cite{peyrard-2019-simple} \\ \cite{fabbri2021summeval}, etc. \end{tabular}               \\ \hline
Transduction                 & Preservation    & Semantic Similarity                                                                              
& \begin{tabular}[c]{@{}l@{}} \cite{mir-etal-2019-evaluating} \\ \cite{yamshchikov2020styletransfer}  \end{tabular}               \\ \hline
\multirow{2}{*}{Creation}    & Engagingness     & \begin{tabular}[c]{@{}l@{}} Depth, \\ (Not) Dull, \\ Interestingness \end{tabular}       
&  \begin{tabular}[c]{@{}l@{}} \cite{venkatesh2018evaluating}, \\ \cite{see2019makes} \\ \cite{mehri2020usr} \\ \cite{Gopalakrishnan2019}, etc.  \end{tabular}               \\ \cline{2-4} 
                             & Groundedness  & \begin{tabular}[c]{@{}l@{}} Persona Distinctiveness \\ Knowledge Usage, \\ Knowledge Injection \end{tabular}                                                                            
&  \begin{tabular}[c]{@{}l@{}}\cite{mehri2020usr} \\ \cite{dinan2019second} \\ \cite{smith-etal-2020-put}, etc. \end{tabular}              \\ \hline
\end{tabular}
\caption{The key aspects discussed for each task category. Examples of prior work that considered each aspect as desirable properties are listed. }
\label{table:task-aspects}
\end{table*}

\clearpage

\onecolumn
\section{Alignment Prediction Example}

\begin{table*}[h]
\centering
\resizebox{\textwidth}{!}{
\begin{tabular}{llllllllll}
\toprule
\multicolumn{10}{l}{\begin{tabular}[c]{@{}l@{}}\textbf{\textsc{Document:}} \textbf{Darth vader} and imperial stormtroopers have invaded a denbighshire seaside town to \\ welcome \textbf{the actor who plays the infamous villain.} \\ Spencer wilding, who \textbf{hails from rhyl}, was \textbf{the guest of honour} at \textbf{a special screening} of rogue one. \\ He had to muster all powers of the force to keep \textbf{his vader role} secret until the film's release. "it's a hell \\ of a secret to keep," said wilding, who was cast as the body \textbf{actor} for the role. \\ "but when you're \textbf{a professional actor} - when you sign that black and white sheet of paper saying \\ you cannot say a word... I'm true to my word and i didn't say anything." \\ Speaking to bbc radio wales' good morning wales programme, the 44-year-old said it proved a \\ tricky task after rumours of the role leaked a year ago. "i've been having hundreds of people every day \\ for a year asking me if i'm \textbf{vader}," he said. "if i had a pound for everyone who asked i'd be buying myself \\ a new death star - and it'd be gold plated." \\ \textbf{The 6ft 7in (2m) tall actor} already has a string of hollywood appearances to his name, including \\ guardians of the galaxy, green   lantern, three harry potter films and   the tv blockbuster game of thrones. \\ He said \textbf{the vader role} came from a regular casting call, first with a self-filmed tape, then a recall to \\ pinewood studios. "it's very, very secretive. We didn't even know exactly what the character was and what \\ film it was until we got there," he said. \\ "i opened up the curtain when i went in the dressing room and there he was - \textbf{vader}. \\ "anybody out there who got into that costume and got \textbf{an audition to be darth vader} alone is very \\ exciting, so to pull the character off as well, it's like 'what!' \\ "i'm always pinching myself - i am definitely awake - it is not a dream, it is just another dream come true." \\ While the \textbf{actor} has the body role, just like his predecessor in the original \textbf{star wars films} david prowse, \\ the voice of \textbf{lord vader} is \textbf{actor} james earl jones. \\ That did not stop wilding trying out the voice during filming. \\ "i'm not james earl jones - nowhere near him - but you know i got close to him i think, which helped the \\ other \textbf{actors} - you know, you've got \textbf{vader} in front of you."\end{tabular}} \\ \hline
\multirow{2}{*}{\begin{tabular}[c]{@{}l@{}}\textbf{\textsc{Summary 1:}}\\ \textbf{\textsc{(BART)}}\end{tabular}}       & A      & \color{red}{welsh} & actor     & who  & plays & darth & vader    & in   & the  \\
                                                                                   & 0.94   & 0.79  & 0.98      & 1.00 & 0.99  & 0.98  & 0.99     & 0.99 & 0.84 \\ \cline{2-10} 
                                                                                   & \color{red}{latest} & star  & wars      & film & has   & been  & honoured & at   & \color{red}{the}  \\
                                                                                   & 0.69   & 0.80  & 0.84      & 0.92 & 0.97  & 0.91  & 0.89     & 0.83 & 0.56 \\ \cline{2-10} 
                                                                                   & \color{red}{london} & \color{red}{film}  & \color{red}{festival.} &      &       &       &          &      &      \\
                                                                                   & 0.47   & 0.56  & 0.63      &      &       &       &          &      &      \\ \hline
\multirow{2}{*}{\begin{tabular}[c]{@{}l@{}}\textbf{\textsc{Summary 2:}}\\ \textbf{\textsc{(Repetition)}}\end{tabular}} & the    & {\color{red} the}   & {\color{red} the}       & {\color{red} the}  & {\color{red} the}   & {\color{red} the}   & {\color{red} the}      & {\color{red} the}  & {\color{red} the}  \\
                                                                                   & 0.83   & 0.61  & 0.56      & 0.53 & 0.49  & 0.48  & 0.50     & 0.53 & 0.57 \\ \cline{2-10} 
                                                                                   & {\color{red} the}    & {\color{red} the}   & {\color{red} the}       & {\color{red} the}  & {\color{red} the}   & {\color{red} the}   &          &      &      \\
                                                                                   & 0.58   & 0.57  & 0.56      & 0.57 & 0.55  & 0.55  &          &      &      \\ \bottomrule
\end{tabular}}
\caption{An example of word-level alignment prediction using discriminative model (\textsc{D}) for an XSUM \cite{narayan-etal-2018-dont} article. \textsc{Summary 1} is generated by BART \cite{lewis-etal-2020-bart} and received a human consistency score of $0$ according to \citet{wang2020asking}, meaning it contains hallucination; \textsc{Summary 2} is a repetition of ``the''. As the predictions show, our model assigns low scores to words in {\color{red}{red}}, which either don't follow directly from the article (``latest'', ``the london film festival'', ``welsh''), or are meaningless repetitions (``the''s).}
\label{table:alignment-example}
\end{table*}

\clearpage

\section{All Summarization Results}

\begin{table*}[h]
\centering
\begin{tabular}{lrrrrrr}
\toprule
{Metric Name} &
  \multicolumn{3}{l}{{Sample-Level Correlations}} &
  \multicolumn{3}{l}{{System-Level Correlations}} \\
\textbf{} &
  \multicolumn{1}{l}{{Pearson}} &
  \multicolumn{1}{l}{{Spearman}} &
  \multicolumn{1}{l}{{Kendall}} &
  \multicolumn{1}{l}{{Pearson}} &
  \multicolumn{1}{l}{{Spearman}} &
  \multicolumn{1}{l}{{Kendall}} \\ \hline
\multicolumn{7}{c}{{Reference-Based Metrics}}                                              \\ \hline
ROUGE-1                  & 0.1811          & 0.1416 & 0.1114 & 0.6648 & 0.7441  & 0.5500          \\
ROUGE-2                  & 0.1583          & 0.1360 & 0.1069 & 0.6610 & 0.7794  & 0.6000          \\
ROUGE-L                  & 0.1578          & 0.1147 & 0.0899 & 0.5180 & 0.0882  & 0.1000          \\
BLEU                     & 0.1794          & 0.1607 & 0.1265 & 0.5872 & 0.0765  & 0.0500          \\
METEOR                   & 0.1832          & 0.1508 & 0.1182 & 0.7157 & 0.8441  & 0.6500          \\
ChrF                     & 0.1750          & 0.1535 & 0.1205 & 0.6446 & 0.8235  & 0.6167          \\
CIDEr                    & 0.0336          & 0.0075 & 0.0058 & 0.0676 & -0.3618 & -0.1833         \\
S3-pyr                   & 0.1624          & 0.1442 & 0.1135 & 0.4616 & 0.6676  & 0.5167          \\
S3-rsp                   & 0.1609          & 0.1490 & 0.1173 & 0.4758 & 0.6647  & 0.5000          \\
SMS                      & 0.2110          & 0.2384 & 0.1876 & 0.7136 & 0.8000  & 0.6000          \\
BERTScore-f              & 0.2030          & 0.1547 & 0.1215 & 0.6318 & 0.0794  & 0.0333          \\
MoverScore               & 0.1899          & 0.1707 & 0.1339 & 0.5616 & 0.0000  & -0.0500         \\ \hline
\multicolumn{7}{c}{{Reference-Free Metrics}}                                               \\ \hline
SummaQA-prob             & 0.1202          & 0.1328 & 0.1045 & 0.7545 & 0.8294  & {0.6667} \\
SummaQA-f                & 0.1572          & 0.1635 & 0.1285 & 0.7198 & 0.8324  & 0.6333          \\
BLANC                    & 0.2183          & 0.2303 & 0.1807 & 0.6294 & 0.7706  & 0.6167          \\
FactCC-prob              & 0.3256          & 0.3410 & 0.2703 & 0.7401 & 0.7990  & 0.6176          \\
SUPERT                   & 0.3665          & 0.3264 & 0.2587 & 0.7274 & 0.7912  & 0.6000          \\ \hline
\multicolumn{7}{c}{{Our Metrics}}                                                          \\ \hline
Ours (E) (BERT-base)     & 0.4359          & 0.3744 & 0.2974 & 0.7886 & 0.7794  & 0.5667          \\
Ours (E) (RoBERTa-large) & 0.3315          & 0.3202 & 0.2518 & 0.6166 & 0.7912  & 0.5833          \\
Ours (D) (CNN/DM) &
  0.5240 &
  \textbf{0.4293} &
  \textbf{0.3422} &
  {0.9146} &
  {0.8324} &
  0.6333 \\
Ours (D) (XSUM)          & \textbf{0.5314} & 0.4273 & 0.3414 & 0.9089 & 0.6059  & 0.3833          \\
Ours (R) (CNN/DM)        & 0.4626          & 0.3871 & 0.3071 & 0.8401 & 0.6000  & 0.4167          \\
Ours (R) (XSUM)          & 0.4868          & 0.3896 & 0.3094 & 0.8473 & 0.5265  & 0.3500          \\ \bottomrule
\end{tabular}
\caption{Correlations of all metrics with Consistency aspect of CNN/DM, using annotations from \cite{fabbri2021summeval}. Reference-based metrics were calculated using 11 references. Our metrics based on the trained alignment models (\texttt{D} and \texttt{R}) both clearly outperform previous metrics.}
\label{table:summeval-consistency}
\end{table*}

\clearpage

\begin{table*}[h]
\centering
\begin{tabular}{lrrrrrr}
\toprule
Metric Name              & \multicolumn{3}{l}{Sample-Level Correlations}      & \multicolumn{3}{l}{System-Level Correlations} \\
 &
  \multicolumn{1}{l}{Pearson} &
  \multicolumn{1}{l}{Spearman} &
  \multicolumn{1}{l}{Kendall} &
  \multicolumn{1}{l}{Pearson} &
  \multicolumn{1}{l}{Spearman} &
  \multicolumn{1}{l}{Kendall} \\ \hline
\multicolumn{7}{c}{Reference-Based Metrics (1 Reference)}                                                                      \\ \hline
ROUGE-1                  & 0.3392          & 0.3337          & 0.2402          & 0.6089        & 0.6000        & 0.4667        \\
ROUGE-2                  & 0.2479          & 0.2581          & 0.1853          & 0.6378        & 0.6176        & 0.4333        \\
ROUGE-L                  & 0.3165          & 0.3153          & 0.2262          & 0.5544        & 0.4059        & 0.2500        \\
BLEU                     & 0.2135          & 0.2565          & 0.1842          & 0.5760        & 0.3912        & 0.2333        \\
METEOR                   & 0.3336          & 0.3268          & 0.2351          & 0.6267        & 0.7176        & 0.5000        \\
ChrF                     & 0.3310          & 0.3305          & 0.2378          & 0.6735        & 0.7559        & 0.5333        \\
CIDEr                    & 0.0457          & -0.0205         & -0.0149         & 0.3348        & 0.2765        & 0.1500        \\
S3-pyr                   & 0.3206          & 0.3087          & 0.2209          & 0.6126        & 0.6824        & 0.4833        \\
S3-rsp                   & 0.2913          & 0.2940          & 0.2107          & 0.6452        & 0.7853        & 0.5667        \\
SMS                      & 0.2461          & 0.2535          & 0.1798          & 0.7681        & 0.7618        & 0.5833        \\
BERTScore-f              & 0.3041          & 0.2937          & 0.2102          & 0.5509        & 0.4206        & 0.2667        \\
MoverScore               & 0.2850          & 0.2898          & 0.2077          & 0.5701        & 0.4735        & 0.3167        \\ \hline
\multicolumn{7}{c}{Reference-Free Metrics}                                                                                     \\ \hline
SummaQA-prob             & 0.1370          & 0.1474          & 0.1039          & 0.6894        & 0.8235        & 0.6333        \\
SummaQA-f                & 0.1665          & 0.1528          & 0.1071          & 0.5217        & 0.4412        & 0.3333        \\
BLANC                    & 0.2552          & 0.2355          & 0.1679          & 0.4690        & 0.3529        & 0.3167        \\
FactCC-prob              & 0.2009          & 0.1576          & 0.1109          & 0.3487        & 0.3162        & 0.2353        \\
SUPERT                   & 0.3282          & 0.2848          & 0.2036          & 0.4569        & 0.3618        & 0.3667        \\ \hline
\multicolumn{7}{c}{Our Metrics (1 Reference)}                                                                                  \\ \hline
Ours (E) (BERT-base)     & 0.3635          & 0.3359          & 0.2401          & 0.6052        & 0.7500        & 0.5833        \\
Ours (E) (RoBERTa-large) & \textbf{0.4985} & \textbf{0.4882} & \textbf{0.3563} & 0.8494        & 0.8412        & 0.7167        \\
Ours (D) (CNN/DM)        & 0.3824          & 0.3499          & 0.2528          & 0.6226        & 0.6000        & 0.4833        \\
Ours (D) (XSUM)          & 0.3802          & 0.3502          & 0.2526          & 0.6126        & 0.5882        & 0.4667        \\
Ours (R) (CNN/DM)        & 0.3733          & 0.3439          & 0.2495          & 0.5556        & 0.4735        & 0.4000        \\
Ours (R) (XSUM)          & 0.3714          & 0.3445          & 0.2493          & 0.5447        & 0.4324        & 0.3667        \\ \bottomrule
\end{tabular}
\caption{Correlations of all considered metrics with Relevance aspect of CNN/DM, using annotations from \cite{fabbri2021summeval}. Reference-based metrics are calculated using 1 reference. Our (XSUM) metrics use $\yv\to\xv$ models based on XSUM alignment data, but still use $\rv\to\yv$ models based on CNN/DM alignment data. Accounting for both reference and input article on top of better alignment modeling, our metrics clearly outperform all other baselines.}
\label{table:summeval-relevance-1}
\end{table*}

\clearpage

\begin{table*}[h]
\centering
\begin{tabular}{lrrrrrr}
\toprule
Metric Name              & \multicolumn{3}{l}{Sample-Level Correlations}      & \multicolumn{3}{l}{System-Level Correlations} \\
 &
  \multicolumn{1}{l}{Pearson} &
  \multicolumn{1}{l}{Spearman} &
  \multicolumn{1}{l}{Kendall} &
  \multicolumn{1}{l}{Pearson} &
  \multicolumn{1}{l}{Spearman} &
  \multicolumn{1}{l}{Kendall} \\ \hline
\multicolumn{7}{c}{Reference-Based Metrics (11 References)}                                                                    \\ \hline
ROUGE-1                  & 0.3565          & 0.3160          & 0.2276          & 0.5486        & 0.7441        & 0.5833        \\
ROUGE-2                  & 0.2685          & 0.2564          & 0.1837          & 0.5659        & 0.6206        & 0.4333        \\
ROUGE-L                  & 0.3347          & 0.2990          & 0.2157          & 0.4364        & 0.3647        & 0.3667        \\
BLEU                     & 0.2750          & 0.2581          & 0.1851          & 0.6189        & 0.5471        & 0.3833        \\
METEOR                   & 0.3237          & 0.2936          & 0.2101          & 0.6217        & 0.7471        & 0.5500        \\
ChrF                     & 0.3561          & 0.3366          & 0.2418          & 0.7017        & 0.7441        & 0.5500        \\
CIDEr                    & -0.0055         & -0.0261         & -0.0191         & 0.1654        & 0.1500        & 0.0833        \\
S3-pyr                   & 0.3469          & 0.3180          & 0.2272          & 0.3811        & 0.2765        & 0.2167        \\
S3-rsp                   & 0.3227          & 0.3101          & 0.2216          & 0.3975        & 0.2971        & 0.2333        \\
SMS                      & 0.2593          & 0.2467          & 0.1765          & 0.5156        & 0.4618        & 0.4000        \\
BERTScore-f              & 0.3192          & 0.2961          & 0.2126          & 0.5991        & 0.5441        & 0.4000        \\
MoverScore               & 0.3114          & 0.3108          & 0.2237          & 0.6419        & 0.5382        & 0.3500        \\ \hline
\multicolumn{7}{c}{Reference-Free Metrics}                                                                                     \\ \hline
SummaQA-prob             & 0.1370          & 0.1474          & 0.1039          & 0.6894        & 0.8235        & 0.6333        \\
SummaQA-f                & 0.1665          & 0.1528          & 0.1071          & 0.5217        & 0.4412        & 0.3333        \\
BLANC                    & 0.2552          & 0.2355          & 0.1679          & 0.4690        & 0.3529        & 0.3167        \\
FactCC-prob              & 0.2009          & 0.1576          & 0.1109          & 0.3487        & 0.3162        & 0.2353        \\
SUPERT                   & 0.3282          & 0.2848          & 0.2036          & 0.4569        & 0.3618        & 0.3667        \\ \hline
\multicolumn{7}{c}{Our Metrics (11 References)}                                                                                \\ \hline
Ours (E) (BERT-base)     & 0.3906          & 0.3547          & 0.2544          & 0.5032        & 0.3765        & 0.3167        \\
Ours (E) (RoBERTa-large) & \textbf{0.5198} & \textbf{0.4990} & \textbf{0.3671} & 0.7539        & 0.7324        & 0.6167        \\
Ours (D) (CNN/DM)        & 0.4423          & 0.3962          & 0.2862          & 0.5821        & 0.4794        & 0.3833        \\
Ours (D) (XSUM)          & 0.4426          & 0.3991          & 0.2878          & 0.5691        & 0.4794        & 0.3833        \\
Ours (R) (CNN/DM)        & 0.4115          & 0.3617          & 0.2644          & 0.4999        & 0.3824        & 0.3333        \\
Ours (R) (XSUM)          & 0.4121          & 0.3680          & 0.2687          & 0.4906        & 0.3353        & 0.2833        \\ \bottomrule
\end{tabular}
\caption{Correlations of all considered metrics with Relevance aspect of CNN/DM, using annotations from \cite{fabbri2021summeval}. Reference-based metrics are calculated using 11 references. Our (XSUM) metrics use $\yv\to\xv$ models based on XSUM alignment data, but still use $\rv\to\yv$ models based on CNN/DM alignment data. Accounting for both reference and input article on top of better alignment modeling, our metrics clearly outperform all other baselines.}
\label{table:summeval-relevance-11}
\end{table*}

\clearpage

\begin{table*}[h]
\centering
\begin{tabular}{lrrrrrr}
\toprule
{Metric Name} &
  \multicolumn{3}{l}{{CNN/DM Correlations}} &
  \multicolumn{3}{l}{{XSUM Correlations}} \\
 &
  \multicolumn{1}{l}{{Pearson}} &
  \multicolumn{1}{l}{{Spearman}} &
  \multicolumn{1}{l}{{Kendall}} &
  \multicolumn{1}{l}{{Pearson}} &
  \multicolumn{1}{l}{{Spearman}} &
  \multicolumn{1}{l}{{Kendall}} \\ \hline
\multicolumn{7}{c}{{Reference-Based Metrics}}                                                                                 \\ \hline
ROUGE-1                  & 0.2874          & -               & -               & 0.1322          & -               & -               \\
ROUGE-2                  & 0.1772          & -               & -               & 0.0895          & -               & -               \\
ROUGE-L                  & 0.2409          & -               & -               & 0.0886          & -               & -               \\
METEOR                   & 0.2665          & -               & -               & 0.1003          & -               & -               \\
BLEU-1                   & 0.2968          & -               & -               & 0.1176          & -               & -               \\
BLEU-2                   & 0.2565          & -               & -               & 0.1168          & -               & -               \\
BLEU-3                   & 0.2396          & -               & -               & 0.0841          & -               & -               \\
BLEU-4                   & 0.2145          & -               & -               & 0.0564          & -               & -               \\
BERTScore-f              & 0.2763          & -               & -               & 0.0251          & -               & -               \\ \hline
\multicolumn{7}{c}{{Reference-Free Metrics}}                                                                                  \\ \hline
FactCC-prob              & 0.4158          & 0.4837          & 0.3758          & 0.2968          & 0.2588          & 0.2118          \\
QAGS                     & 0.5453          & -               & -               & 0.1749          & -               & -               \\ \hline
\multicolumn{7}{c}{{Our Metrics}}                                                                                             \\ \hline
Ours (E) (BERT-base)     & 0.6083          & 0.5180          & 0.4074          & 0.1436          & 0.1437          & 0.1176          \\
Ours (E) (RoBERTa-large) & 0.6091          & 0.5229          & 0.4141          & 0.0548          & 0.0489          & 0.0400          \\
Ours (D) (CNN/DM)        & 0.6188          & \textbf{0.5640} & \textbf{0.4500} & 0.3085          & 0.2952          & 0.2416          \\
Ours (D) (XSUM)          & 0.6205          & 0.5362          & 0.4260          & \textbf{0.3222} & \textbf{0.3149} & \textbf{0.2576} \\
Ours (R) (CNN/DM)        & 0.6468          & 0.5252          & 0.4180          & 0.2157          & 0.1949          & 0.1594          \\
Ours (R) (XSUM)          & \textbf{0.6612} & 0.5445          & 0.4348          & 0.2718          & 0.2509          & 0.2053          \\ \bottomrule
\end{tabular}
\caption{Sample-Level correlations of all considered metrics with Consistency aspect of CNN/DM and XSUM, based on annotations from \cite{wang2020asking}. Spearman and Kendall-Tau correlations for baseline metrics were not reported, except for \texttt{FactCC} which we computed on our own. System-level correlations are not reported due to dataset limits. On CNN/DM, all of our metrics outperform previous metrics. On XSUM, our \texttt{D}-based metric trained on the same domain also achieves the best performance. The \texttt{E}-based metrics see a catastrophic drop likely due to the higher abstractiveness of XSUM that renders embedding matching inadequate. }
\label{table:qags-consistency}
\end{table*}

\clearpage

\section{All Style Transfer Results}

\begin{table*}[h]
\centering
\begin{tabular}{lrrrrrr}
\toprule
{Metric Name} &
  \multicolumn{3}{l}{{Sample-Level Correlations}} &
  \multicolumn{3}{l}{{System-Level Correlations}} \\
 &
  \multicolumn{1}{l}{{Pearson}} &
  \multicolumn{1}{l}{{Spearman}} &
  \multicolumn{1}{l}{{Kendall}} &
  \multicolumn{1}{l}{{Pearson}} &
  \multicolumn{1}{l}{{Spearman}} &
  \multicolumn{1}{l}{{Kendall}} \\ \hline
\multicolumn{7}{c}{{Lexical-Matching Based Metrics}}                              \\ \hline
BLEU                     & 0.4361 & 0.4303 & 0.3125 & 0.8476 & 0.8042 & 0.6364 \\
ROUGE-1                  & 0.4808 & 0.4981 & 0.3614 & 0.7749 & 0.6503 & 0.4545 \\
ROUGE-2                  & 0.4664 & 0.4589 & 0.3421 & 0.8444 & 0.7622 & 0.5455 \\
ROUGE-L                  & 0.4833 & 0.4975 & 0.3606 & 0.7814 & 0.7133 & 0.5152 \\
METEOR                   & 0.4764 & 0.4993 & 0.3592 & 0.8291 & 0.7622 & 0.5455 \\
ChrF                     & 0.5048 & 0.5028 & 0.3632 & 0.8026 & 0.7343 & 0.5758 \\ \hline
\multicolumn{7}{c}{{Embedding Based Metrics}}                           \\ \hline
EmbedAvg                 & 0.3248 & 0.4127 & 0.2959 & 0.7272 & 0.5944 & 0.3939 \\
GreedyMatch              & 0.4542 & 0.4760 & 0.3434 & 0.7431 & 0.6084 & 0.4242 \\
VectorExtrema            & 0.4571 & 0.4589 & 0.3306 & 0.7702 & 0.6364 & 0.4242 \\
WMD                      & 0.4902 & 0.5047 & 0.3646 & 0.7776 & 0.6713 & 0.5455 \\ \hline
\multicolumn{7}{c}{{Pre-Trained Model Based Metrics}}                   \\ \hline
BERTScore                & 0.5185 & 0.5187 & 0.3751 & 0.8078 & 0.7133 & 0.5152 \\
MoverScore               & 0.5209 & 0.5148 & 0.3734 & 0.8308 & 0.7622 & 0.5455 \\
BLEURT                   & 0.5043 & 0.4934 & 0.3566 & 0.8673 & 0.7902 & 0.6061 \\ \hline
\multicolumn{7}{c}{{Our Metrics}}                                       \\ \hline
Ours (E) (BERT-base)     & 0.5147 & 0.5169 & 0.3740 & 0.8096 & 0.7133 & 0.5152 \\
Ours (E) (RoBERTa-large) & 0.5142 & 0.5150 & 0.3752 & 0.8618 & 0.7832 & 0.5758 \\
\begin{tabular}[c]{@{}l@{}}Ours (E) \\ (RoBERTa-large-MNLI-9)\end{tabular}  &
  \textbf{0.5216} &
  \textbf{0.5236} &
  \textbf{0.3805} &
  0.8081 &
  0.7133 &
  0.5152 \\
Ours (D)                 & 0.4974 & 0.4952 & 0.3579 & 0.8385 & 0.7483 & 0.5152 \\
Ours (R)                 & 0.5060 & 0.5059 & 0.3645 & 0.8226 & 0.6993 & 0.4848 \\ \bottomrule
\end{tabular}
\caption{Correlations of all considered metrics with Preservation aspect of Yelp, using annotations from \cite{mir-etal-2019-evaluating}. Explicitly accounting for two-way input-out alignments in an ``F1''-style harmonic mean aggregation (Eq.\ref{eq:metric:preservation}), our metrics (\texttt{E}) achieve competitive or better performance than previous metrics. Our \texttt{D}- and \texttt{R}-based metrics fall behind slightly, likely because one-to-one token matching is more suitable for two text pieces with similar information volume.}
\label{table:yelp-preservation}
\end{table*}

\clearpage

\section{All Dialog Results}
\label{sec:appendix:dialog}

\begin{table*}[h]
\centering
\begin{tabular}{lrrrrrr}
\toprule
{Metric Name} &
  \multicolumn{3}{l}{{Sample-Level Correlations}} &
  \multicolumn{3}{l}{{System-Level Correlations}} \\
 &
  \multicolumn{1}{l}{{Pearson}} &
  \multicolumn{1}{l}{{Spearman}} &
  \multicolumn{1}{l}{{Kendall}} &
  \multicolumn{1}{l}{{Pearson}} &
  \multicolumn{1}{l}{{Spearman}} &
  \multicolumn{1}{l}{{Kendall}} \\ \hline
\multicolumn{7}{c}{{Reference Based Metrics}}                                                        \\ \hline
F-1                      & 0.0473          & 0.0132          & -               & 0.9956  & 1.0000  & -      \\
BLEU-1                   & -0.1081         & -0.0922         & -               & 0.2599  & 0.6000  & -      \\
BLEU-2                   & -0.1048         & -0.1010         & -               & 0.6816  & 0.4000  & -      \\
BLEU-3                   & -0.1247         & -0.1151         & -               & 0.6668  & 0.4000  & -      \\
BLEU-4                   & -0.1359         & -0.1242         & -               & 0.8413  & 0.8000  & -      \\
METEOR                   & -0.0458         & 0.0116          & -               & 0.9065  & 0.8000  & -      \\
ROUGE-L                  & -0.1456         & -0.1354         & -               & 0.1710  & 0.0000  & -      \\
BERTScore (base)         & 0.0325          & 0.0491          & -               & 0.5173  & 0.8000  & -      \\
BERTScore (large)        & -0.0418         & -0.0245         & -               & 0.2410  & 0.0000  & -      \\ \hline
\multicolumn{7}{c}{{Reference-Free Metrics}}                                                         \\ \hline
FED-Interesting          & 0.1818          & 0.0255          & 0.0182          & 0.9277  & 1.0000  & 1.0000 \\
USR-MLM                  & -0.1045         & -0.1007         & -               & -0.2842 & -0.4000 & -      \\
USR-DR (x=c)             & 0.0606          & 0.2634          & -               & 0.8202  & 1.0000  & -      \\
USR-DR (x=f)             & -0.0022         & -0.0039         & -               & -0.0178 & -0.2108 & -      \\
USR                      & 0.0315          & 0.0171          &  -              & 0.8084  & 1.0000  & -      \\
Word Length              & 0.3910          & 0.4220          & 0.3267          & 0.8965  & 0.8000  & 0.6000 \\ \hline
\multicolumn{7}{c}{{Our Metrics}}                                                                    \\ \hline
Ours (E) (BERT-base)     & 0.5003          & 0.5490          & 0.4193          & 0.9061  & 1.0000  & 1.0000 \\
Ours (E) (RoBERTa-large) & 0.4081          & 0.4502          & 0.3375          & 0.9003  & 0.9000  & 0.8000 \\
Ours (D) (PersonaChat)   & 0.5265          & {0.5793} & \textbf{0.4412} & 0.9425  & 1.0000  & 1.0000 \\
Ours (D) (TopicalChat)   & 0.5317          & \textbf{0.5818}          & 0.4409          & 0.9447  & 1.0000  & 1.0000 \\
Ours (R) (PersonaChat)   & \textbf{0.5320} & 0.5692          & 0.4346          & 0.9433  & 1.0000  & 1.0000 \\
Ours (R) (TopicalChat)   & 0.4933          & 0.5333          & 0.4043          & 0.9244  & 1.0000  & 1.0000 \\ \bottomrule
\end{tabular}
\caption{Correlations of all considered metrics with Engagingness aspect of PersonaChat, using annotations from \cite{mehri2020usr}. Kendall-Tau correlations for baseline metrics were not reported. By measuring aligned information volume ($\mathrm{sum}$) and with accurate estimation models, our metrics with different implementations all improve over previous methods by large margins.}
\label{table:personachat-engagingness}
\end{table*}

\clearpage

\begin{table*}[h]
\centering
\begin{tabular}{lrrrrrr}
\toprule
{Metric Name} &
  \multicolumn{3}{l}{{Sample-Level Correlations}} &
  \multicolumn{3}{l}{{System-Level Correlations}} \\
 &
  \multicolumn{1}{l}{{Pearson}} &
  \multicolumn{1}{l}{{Spearman}} &
  \multicolumn{1}{l}{{Kendall}} &
  \multicolumn{1}{l}{{Pearson}} &
  \multicolumn{1}{l}{{Spearman}} &
  \multicolumn{1}{l}{{Kendall}} \\ \hline
\multicolumn{7}{c}{{Reference Based Metrics}}                                \\ \hline
F-1                      & 0.0869  & 0.1056  & -       & 0.9956  & 1.0000  & -      \\
BLEU-1                   & 0.0737  & 0.0729  & -       & 0.2599  & 0.6000  & -      \\
BLEU-2                   & 0.1083  & 0.0722  & -       & 0.6816  & 0.4000  & -      \\
BLEU-3                   & 0.0999  & 0.0594  & -       & 0.6668  & 0.4000  & -      \\
BLEU-4                   & 0.0698  & 0.0528  & -       & 0.8413  & 0.8000  & -      \\
METEOR                   & 0.1678  & 0.1719  & -       & 0.9065  & 0.8000  & -      \\
ROUGE-L                  & 0.0710  & 0.0632  & -       & 0.1710  & 0.0000  & -      \\
BERTScore (base)         & 0.0719  & 0.0465  & -       & 0.5173  & 0.8000  & -      \\
BERTScore (large)        & 0.0271  & 0.0094  & -       & 0.2410  & 0.0000  & -      \\ \hline
\multicolumn{7}{c}{{Reference-Free Metrics}}                                 \\ \hline
FED-Informative          & 0.0165  & -0.0405 & -0.0315 & 0.9015  & 0.9000  & 0.8000 \\
USR-MLM                  & -0.0782 & -0.0756 & -       & -0.2842 & -0.4000 & -      \\
USR-DR (x=c)             & 0.4508  & 0.6309  & -       & 0.8202  & 1.0000  & -      \\
USR-DR (x=f)             & -0.0927 & -0.0903 & -       & -0.0178 & -0.2108 & -      \\
USR                      & 0.4027  & 0.3177  & -       & 0.8084  & 1.0000  & -      \\ 
Word Length              & 0.3171  & 0.3051  & 0.2467  & 0.7698  & 0.5000  & 0.4000 \\ \hline
\multicolumn{7}{c}{{Our Metrics}}                                            \\ \hline
Ours (E) (BERT-base)     & 0.5761  & 0.5683  & 0.4492  & 0.8720  & 0.9000  & 0.8000 \\
Ours (E) (RoBERTa-large) & 0.3758  & 0.3652  & 0.2862  & 0.8140  & 0.7000  & 0.6000 \\
Ours (D) (PersonaChat)   & 0.5683  & 0.5674  & 0.4505  & 0.8684  & 0.9000  & 0.8000 \\
Ours (D) (TopicalChat)   & 0.4056  & 0.4172  & 0.3270  & 0.7536  & 0.5000  & 0.4000 \\
Ours (R) (PersonaChat)   & 0.6597  & 0.6689  & 0.5338  & 0.9151  & 1.0000  & 1.0000 \\
Ours (R) (TopicalChat) &
  \textbf{0.6819} &
  \textbf{0.7113} &
  \textbf{0.5636} &
  0.9420 &
  1.0000 &
  1.0000 \\ \bottomrule
\end{tabular}
\caption{Correlations of all considered metrics with Groundedness aspect of PersonaChat, using annotations from \cite{mehri2020usr}. Kendall-Tau correlations for baseline metrics were not reported. Trained with aggregated alignment scores and benefiting from the expressivity of end-to-end models, our regression-based (\texttt{R}) metrics strongly outperform all other metrics.}
\label{table:personachat-groundedness}
\end{table*}

\clearpage

\begin{table*}[h]
\centering
\begin{tabular}{lrrrrrr}
\toprule
{Metric Name} &
  \multicolumn{3}{l}{{Sample-Level Correlations}} &
  \multicolumn{3}{l}{{System-Level Correlations}} \\
 &
  \multicolumn{1}{l}{{Pearson}} &
  \multicolumn{1}{l}{{Spearman}} &
  \multicolumn{1}{l}{{Kendall}} &
  \multicolumn{1}{l}{{Pearson}} &
  \multicolumn{1}{l}{{Spearman}} &
  \multicolumn{1}{l}{{Kendall}} \\ \hline
\multicolumn{7}{c}{{Reference Based Metrics}}                                                                         \\ \hline
F-1                      & 0.2523          & 0.2565          & -                    & 0.5944 & 0.6000 & -                    \\
BLEU-1                   & 0.3144          & 0.3343          & -                    & 0.8197 & 0.7000 & -                    \\
BLEU-2                   & 0.3184          & 0.3323          & -                    & 0.8099 & 0.9000 & -                    \\
BLEU-3                   & 0.2782          & 0.3247          & -                    & 0.9047 & 0.9000 & -                    \\
BLEU-4                   & 0.2322          & 0.3156          & -                    & 0.8883 & 0.9000 & -                    \\
METEOR                   & 0.3668          & 0.4391          & -                    & 0.9398 & 0.9000 & -                    \\
ROUGE-L                  & 0.2946          & 0.2995          & -                    & 0.8084 & 0.9000 & -                    \\
BERTScore (base)         & 0.3512          & 0.3725          & -                    & 0.9108 & 0.9000 & -                    \\
BERTScore (large)        & 0.3167          & 0.3349          & -                    & 0.8480 & 0.9000 & -                    \\ \hline
\multicolumn{7}{c}{{Reference-Free Metrics}}                                                                          \\ \hline
FED-Interesting          & -0.0004         & -0.0328         & -0.0230 & 0.8881 & 0.8286 & 0.7333 \\
USR-MLM                  & 0.3189          & 0.3337          & -                    & 0.4663 & 0.9000 & -                    \\
USR-DR (x=c)             & 0.3533          & 0.4877          & -                    & 0.9233 & 0.7000 & -                    \\
USR-DR (x=f)             & 0.2006          & 0.4110          & -                    & 0.8685 & 0.9000 & -                    \\
USR                      & 0.4555          & 0.4645          & -                    & 0.9297 & 1.0000 & -                    \\
Word Length              & 0.4079          & 0.4089          & 0.3053               & 0.9662 & 0.8286 & 0.7333               \\ \hline
\multicolumn{7}{c}{{Our Metrics}}                                                                                     \\ \hline
Ours (E) (BERT-base)     & 0.4937          & 0.5047          & 0.3710               & 0.9606 & 0.9429 & 0.8667               \\
Ours (E) (RoBERTa-large) & 0.4471          & 0.4479          & 0.3288               & 0.9617 & 0.8286 & 0.7333               \\
Ours (D) (PersonaChat)   & 0.5124          & 0.5245          & 0.3878               & 0.9572 & 0.6571 & 0.6000               \\
Ours (D) (TopicalChat)   & \textbf{0.5163} & \textbf{0.5253} & \textbf{0.3873}      & 0.9657 & 0.8286 & 0.7333               \\
Ours (R) (PersonaChat)   & 0.4542          & 0.4588          & 0.3357               & 0.9529 & 0.8286 & 0.7333               \\
Ours (R) (TopicalChat)   & 0.4653          & 0.4643          & 0.3395               & 0.9492 & 0.8286 & 0.7333               \\ \bottomrule
\end{tabular}
\caption{Correlations of all considered metrics with Engagingness aspect of TopicalChat, using annotations from \cite{mehri2020usr}. Kendall-Tau correlations for baseline metrics were not reported. By measuring aligned information volume ($\mathrm{sum}$) and with accurate estimation models, our metrics with different implementations all compete with or improve over previous methods.}
\label{table:topicalchat-engagingness}
\end{table*}

\clearpage

\begin{table*}[h]
\centering
\begin{tabular}{lrrrrrr}
\toprule
{Metric Name} &
  \multicolumn{3}{l}{{Sample-Level Correlations}} &
  \multicolumn{3}{l}{{System-Level Correlations}} \\
 &
  \multicolumn{1}{l}{{Pearson}} &
  \multicolumn{1}{l}{{Spearman}} &
  \multicolumn{1}{l}{{Kendall}} &
  \multicolumn{1}{l}{{Pearson}} &
  \multicolumn{1}{l}{{Spearman}} &
  \multicolumn{1}{l}{{Kendall}} \\ \hline
\multicolumn{7}{c}{{Reference Based Metrics}}                           \\ \hline
F-1                      & 0.1495 & 0.1485 & -      & 0.5970 & 0.6000 & -      \\
BLEU-1                   & 0.2888 & 0.3033 & -      & 0.8357 & 0.7000 & -      \\
BLEU-2                   & 0.2819 & 0.3066 & -      & 0.8309 & 0.9000 & -      \\
BLEU-3                   & 0.2442 & 0.3106 & -      & 0.9259 & 0.9000 & -      \\
BLEU-4                   & 0.2126 & 0.3096 & -      & 0.9084 & 0.9000 & -      \\
METEOR                   & 0.3328 & 0.3909 & -      & 0.9534 & 0.9000 & -      \\
ROUGE-L                  & 0.3099 & 0.3273 & -      & 0.8333 & 0.9000 & -      \\
BERTScore (base)         & 0.2847 & 0.2947 & -      & 0.9308 & 0.9000 & -      \\
BERTScore (large)        & 0.2909 & 0.3167 & -      & 0.8703 & 0.9000 & -      \\ \hline
\multicolumn{7}{c}{{Reference-Free Metrics}}                            \\ \hline
FED-Informative          & 0.0311 & 0.0243 & 0.0209 & 0.9340 & 0.7143 & 0.6000 \\
USR-MLM                  & 0.2195 & 0.2261 & -      & 0.5070 & 0.9000 & -      \\
USR-DR (x=c)             & 0.2285 & 0.4179 & -      & 0.9155 & 0.7000 & -      \\
USR-DR (x=f)             & 0.2220 & 0.4468 & -      & 0.8884 & 0.9000 & -      \\
USR                      & 0.3175 & 0.3353 & -      & 0.9469 & 1.0000 & -      \\
Word Length              & 0.2624 & 0.2681 & 0.2084 & 0.9859 & 0.8286 & 0.7333 \\ \hline
\multicolumn{7}{c}{{Our Metrics}}                                       \\ \hline
Ours (E) (BERT-base)     & 0.4293 & 0.3949 & 0.3075 & 0.9784 & 0.9429 & 0.8667 \\
Ours (E) (RoBERTa-large) & 0.3122 & 0.3141 & 0.2421 & 0.9868 & 0.8286 & 0.7333 \\
Ours (D) (PersonaChat)   & 0.3697 & 0.3691 & 0.2856 & 0.9784 & 0.9429 & 0.8667 \\
Ours (D) (TopicalChat)   & 0.3099 & 0.3159 & 0.2421 & 0.9842 & 0.8286 & 0.7333 \\
Ours (R) (PersonaChat)   & 0.4026 & 0.3788 & 0.3137 & 0.9742 & 0.7143 & 0.6    \\
Ours (R) (TopicalChat) &
  \textbf{0.5235} &
  \textbf{0.4768} &
  \textbf{0.3838} &
  0.9674 &
  0.8857 &
  0.7333 \\ \bottomrule
\end{tabular}
\caption{Correlations of all considered metrics with Groundedness aspect of TopicalChat, using annotations from \cite{mehri2020usr}. Kendall-Tau correlations for baseline metrics were not reported. Trained with aggregated alignment scores and benefiting from the expressivity of end-to-end models, our regression-based (\texttt{R}) metric trained on TopicalChat strongly outperforms all other metrics.}
\label{table:topicalchat-groundedness}
\end{table*}

\clearpage

\begin{table}[t]
\centering
\begin{tabular}{lllll}
\toprule
Metric           & \multicolumn{2}{l}{PersonaChat}   & \multicolumn{2}{l}{TopicalChat}   \\
 & \multicolumn{1}{l}{Swapped} & \multicolumn{1}{l}{Ours} & \multicolumn{1}{l}{Swapped} & \multicolumn{1}{l}{Ours} \\ \hline
Engagingness (E) & \textbf{0.5082} & 0.5003          & \textbf{0.5180} & 0.4973          \\
Engagingness (D) & 0.4725          & \textbf{0.5265} & 0.4708          & \textbf{0.5163} \\
Engagingness (R) & 0.4679          & \textbf{0.5320} & \textbf{0.4902} & 0.4653          \\
Groundedness (E) & 0.5323          & \textbf{0.5761} & 0.3870          & \textbf{0.4293} \\
Groundedness (D) & 0.4945          & \textbf{0.5683} & \textbf{0.3752} & 0.3099          \\
Groundedness (R) & 0.4798          & \textbf{0.6597} & 0.3237          & \textbf{0.5235} \\ \bottomrule
\end{tabular}
\caption{Ablation Studies: Pearson correlations with engagingness and groundedness for dialog tasks with swapped formulas vs our definition. By swapping, we use our engagingness metric to measure groundedness, and vice versa. PersonalChat swaps see across-the-board decreases in correlations, indicating the importance of using our designed formulas on this dataset. TopicalChat swaps see correlation increases more frequently, but the best methods still retain their edge.}
\label{table:dialog-formula}
\end{table}

\begin{table*}[h]
\centering
\begin{tabular}{lrrrrrrrr}
\toprule
Metric Name      & \multicolumn{4}{l}{PersonaChat}                       & \multicolumn{4}{l}{TopicalChat}                      \\
 &
  \multicolumn{1}{l}{U} &
  \multicolumn{1}{l}{N} &
  \multicolumn{1}{l}{MC} &
  \multicolumn{1}{l}{O} &
  \multicolumn{1}{l}{U} &
  \multicolumn{1}{l}{N} &
  \multicolumn{1}{l}{MC} &
  \multicolumn{1}{l}{O} \\ \hline
\multicolumn{9}{c}{Reference-Based Metrics}                                                                                     \\ \hline
F-1              & -0.0340 & 0.0815  & 0.1073          & 0.1422          & 0.0425 & 0.0301  & 0.1290          & 0.1645          \\
BLEU-4           & 0.0537  & 0.1081  & 0.1467          & 0.1353          & 0.2010 & 0.1799  & 0.1307          & 0.2160          \\
METEOR           & 0.0820  & 0.0989  & 0.2500          & 0.2527          & 0.2452 & 0.2121  & 0.2495          & 0.3365          \\
ROUGE-L          & 0.0346  & 0.0096  & 0.1135          & 0.0659          & 0.2069 & 0.1760  & 0.1928          & 0.2745          \\
BERTScore (base) & 0.0676  & 0.0606  & 0.1770          & 0.1690          & 0.2611 & 0.2164  & 0.2432          & 0.3229          \\ \hline
\multicolumn{9}{c}{Reference-Free Metrics}                                                                                      \\ \hline
FED              & 0.0314  & 0.0870  & -0.0634         & -0.0786         & 0.0469 & 0.0482  & -0.1915         & -0.1393         \\
USR-MLM          & 0.1313  & 0.0999  & 0.1805          & 0.0795          & 0.3264 & 0.3370  & 0.3099          & 0.3086          \\
USR-DR (x=c)     & 0.0728  & 0.1733  & 0.6021          & \textbf{0.4814} & 0.1500 & 0.1325  & 0.3391          & 0.3245          \\
USR-DR (x=f)     & -0.0390 & -0.0033 & -0.0198         & -0.0495         & 0.0881 & 0.0313  & 0.0594          & 0.1419          \\
USR              & 0.0997  & 0.1862  & \textbf{0.6065} & 0.4693          & 0.2932 & 0.2260  & \textbf{0.4160} & \textbf{0.4192} \\
Response Length  & -0.0525 & -0.0342 & 0.0901          & 0.2526          & 0.0845 & 0.1253  & 0.2458          & 0.3343          \\ \hline
\multicolumn{9}{c}{Our Metrics}                                                                                                 \\ \hline
Ours (E)         & 0.1185  & 0.1891  & 0.2786          & 0.3690          & 0.2168 & 0.1528  & 0.1669          & 0.2483          \\
Ours (D) &
  \textbf{0.1421} &
  \textbf{0.3384} &
  0.3837 &
  0.4500 &
  \textbf{0.3433} &
  \textbf{0.3653} &
  0.2969 &
  0.3620 \\
Ours (R)         & 0.0639  & 0.1595  & 0.2518          & 0.2076          & 0.0105 & -0.0524 & -0.0248         & -0.0338         \\ \bottomrule
\end{tabular}
\caption{Sample-level Pearson correlations for the remaining aspects in the annotations of \cite{mehri2020usr}, including understandable (U), natural (N), maintains context (MC) and overall (O). Our metric here is the average alignment confidence from response $\yv$ to dialogue history $\xv$ and knowledge $\cv$, i.e. $\mathrm{mean}\left(\textit{align}(\yv\to[\xv,\cv]\right)$, which outperforms existing metrics on \textit{understandability} and \textit{naturalness}.}
\label{table:additional_correlations}
\end{table*}


\end{document}